%% file: main.tex
\documentclass[lettersize,journal]{IEEEtran}
\usepackage{amsmath,amsfonts}
\usepackage{algorithmic}
\usepackage{algorithm}
\usepackage{array}
\usepackage[caption=false,font=normalsize,labelfont=sf,textfont=sf]{subfig}
\usepackage{textcomp}
\usepackage{stfloats}
\usepackage{url}
\usepackage{verbatim}
\usepackage{graphicx}
\usepackage{cite}
\usepackage{tabularx}
\usepackage{multirow}
\usepackage{xcolor}         
\usepackage{graphicx}
\usepackage{bm}
\usepackage{hyperref}
\usepackage{cleveref}
\usepackage{booktabs}
\newcommand{\etal}{\textit{et al.}}

\definecolor{MyGreen}{RGB}{0, 128, 0}

\begin{document}

\title{GaussNav: Gaussian Splatting for Visual Navigation}

\author{Xiaohan Lei, Min Wang, Wengang Zhou,~\IEEEmembership{Senior Member,~IEEE}, Houqiang Li,~\IEEEmembership{Fellow,~IEEE}
\thanks{
This work was supported by National Natural Science Foundation of China
under Contract 62472141, Key Laboratory of Target Cognition and Application Technology under Contract 2023-CXPT-LC-005, and the Youth Innovation Promotion Association CAS. It was also supported by the GPU cluster built by MCC Lab of Information Science and Technology Institution, USTC, and the Supercomputing Center of the USTC. (\textit{Corresponding authors: Min Wang and Wengang Zhou})

Xiaohan Lei, Wengang Zhou, and Houqiang Li are with the MoE Key Laboratory of Brain-inspired Intelligent Perception and Cognition, University of Science and Technology of China, Hefei, 230027, China (e-mail: leixh@mail.ustc.edu.cn; zhwg@ustc.edu.cn; lihq@ustc.edu.cn).
Min Wang is with the Institute of Artificial Intelligence, Hefei Comprehensive National Science Center, Hefei 230030, China (e-mail: wangmin@iai.ustc.edu.cn).}
}

\markboth{IEEE TRANSACTIONS ON PATTERN ANALYSIS AND MACHINE INTELLIGENCE}
{IEEE TRANSACTIONS ON PATTERN ANALYSIS AND MACHINE INTELLIGENCE}



\maketitle

\begin{abstract}
In embodied vision, Instance ImageGoal Navigation (IIN) requires an agent to locate a specific object depicted in a goal image within an unexplored environment. The primary challenge of IIN arises from the need to recognize the target object across varying viewpoints while ignoring potential distractors. Existing map-based navigation methods typically use Bird's Eye View (BEV) maps, which lack detailed texture representation of a scene. Consequently, while BEV maps are effective for semantic-level visual navigation, they are struggling for instance-level tasks.
To this end, we propose a new framework for IIN, Gaussian Splatting for Visual Navigation (GaussNav), which constructs a novel map representation based on 3D Gaussian Splatting (3DGS). The GaussNav framework enables the agent to memorize both the geometry and semantic information of the scene, as well as retain the textural features of objects. 
By matching renderings of similar objects with the target, the agent can accurately identify, ground, and navigate to the specified object.
Our GaussNav framework demonstrates a significant performance improvement, with Success weighted by Path Length (SPL) increasing from 0.347 to 0.578 on the challenging Habitat-Matterport 3D (HM3D) dataset. The source code is publicly available at the link: \url{https://github.com/XiaohanLei/GaussNav}.
\end{abstract}


\input{sections/intruduction}

\input{sections/related_work}

\input{sections/methods}

\input{sections/experiments}

\input{sections/conclusion}

{\small
\bibliographystyle{IEEEtran}
\bibliography{main}
}

\begin{IEEEbiography}[{\includegraphics[width=1in,height=1.25in,clip,keepaspectratio]{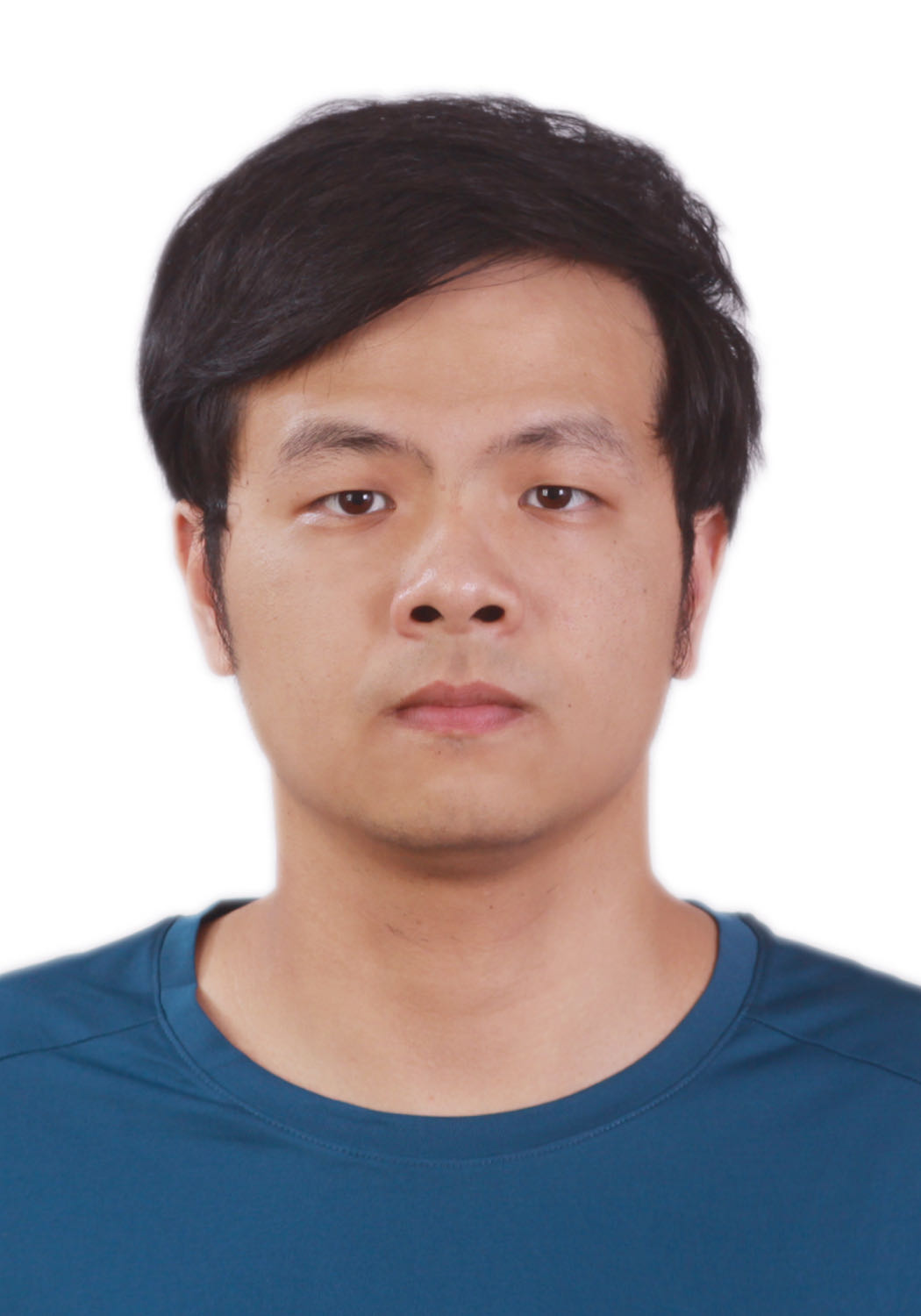}}]{Xiaohan Lei} is currently pursuing the Ph.D. degree in information and communication engineering with the Department of Information Science and Technology, from the University of Science and Technology of China. 

His research interests include embodied visual navigation, robot manipulation and embodied computer vision.
\end{IEEEbiography}

\begin{IEEEbiography}[{\includegraphics[width=1in,height=1.25in,clip,keepaspectratio]{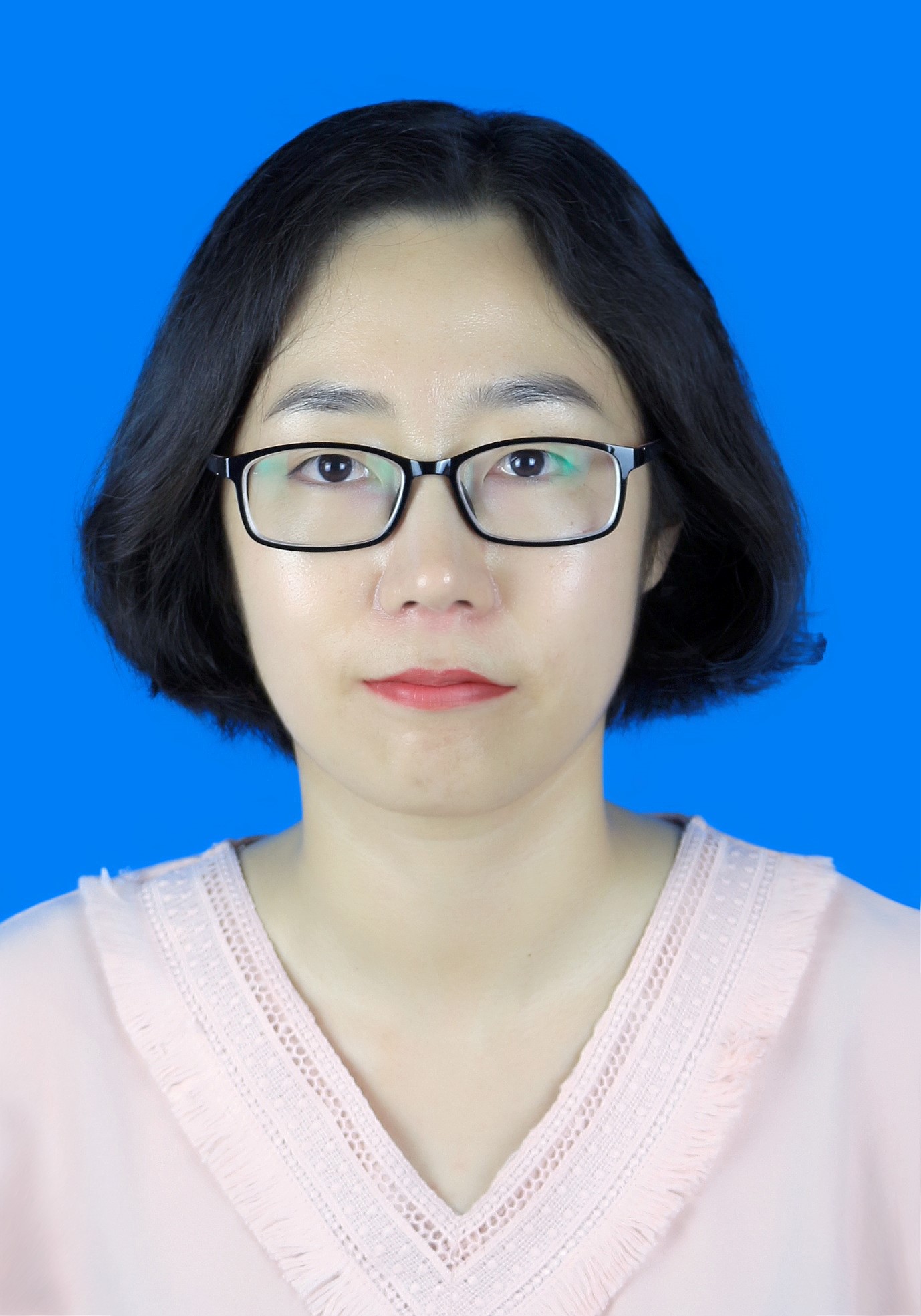}}]{Min Wang} received the B.E., and Ph.D degrees in electronic information engineering from University of Science and Technology of China (USTC), in 2014 and 2019, respectively. She is working in Institute of Artificial Intelligence, Hefei Comprehensive National Science Center. 

Her current research interests include binary hashing, multimedia information retrieval and computer vision.
\end{IEEEbiography}

\begin{IEEEbiography}[{\includegraphics[width=1in,height=1.25in,clip,keepaspectratio]{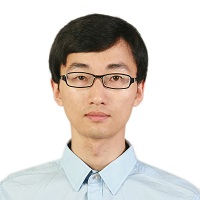}}]{Wengang Zhou} received the B.E. degree in electronic information engineering from Wuhan University, China, in 2006, and the Ph.D. degree in electronic engineering and information science from the University of Science and Technology of China (USTC), China, in 2011. From September 2011 to September 2013, he worked as a postdoc researcher in Computer Science Department at the University of Texas at San Antonio. He is currently a Professor at the EEIS Department, USTC. 

His research interests include multimedia infor-mation retrieval, computer vision, and computer game. In those fields, he has published over 100 papers in IEEE/ACM Transactions and CCF Tier-A International Conferences. He is the winner of National Science Funds of China (NSFC) for Excellent Young Scientists in 2018, and Chinese Society of Image and Graphics (CSIG) Young Scientist Award in 2024. He is the recipient of the Best Paper Award for ICIMCS 2012. He received the award for the Excellent Ph.D Supervisor of Chinese Society of Image and Graphics (CSIG) in 2021, and the award for the Excellent Ph.D Supervisor of Chinese Academy of Sciences (CAS) in 2022. He won the First Class Wu-Wenjun Award for Progress in Artificial Intelligence Technology in 2021. He served as the publication chair of IEEE ICME 2021 and won 2021 ICME Outstanding Service Award. He is currently an Associate Editor and a Lead Guest Editor of IEEE Transactions on Multimedia, and is the recipient of 2023 IEEE Transactions on Multimedia (TMM) Excellent Editor Award. 
\end{IEEEbiography}

\begin{IEEEbiography}[{\includegraphics[width=1in,height=1.25in,clip,keepaspectratio]{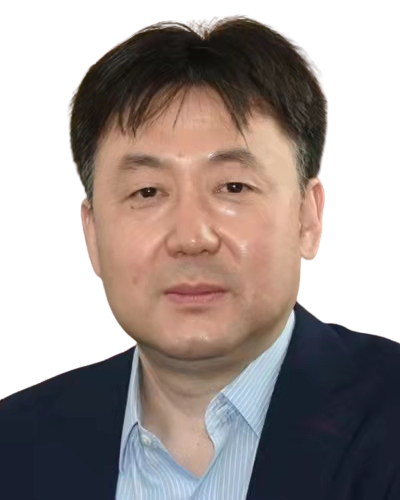}}]{Houqiang Li} (S’12, F’21) received the B.S., M.Eng., and Ph.D. degrees in electronic engineering from the University of Science and Technology of China, Hefei, China, in 1992, 1997, and 2000, respectively. He was elected as a Fellow of IEEE (2021) and he is currently a Professor with the Department of Electronic Engineering and Information Science. 

His research interests include reinforcement learning, multimedia search, image/video analysis, video coding and communication, etc. He has authored and co-authored over 200 papers in journals and conferences. He is the winner of National Science Funds (NSFC) for Distinguished Young Scientists, the Distinguished Professor of Changjiang Scholars Program of China, and the Leading Scientist of Ten Thousand Talent Program of China. He is the associate editor (AE) of IEEE TMM and served as the AE of IEEE TCSVT. He served as the General Co-Chair of ICME 2021 and the TPC Co-Chair of VCIP 2010. He was the recipient of National Technological Invention Award of China (second class) in 2019 and the recipient of National Natural Science Award of China (second class) in 2015. He was the recipient of the Best Paper Award for VCIP 2012, the recipient of the Best Paper Award for ICIMCS 2012, and the recipient of the Best Paper Award for ACM MUM in 2011.
\end{IEEEbiography}

\end{document}

%% file: sections/intruduction.tex
\section{Introduction}

\begin{figure}[tb]
  \centering
  \includegraphics[width=0.9\linewidth]{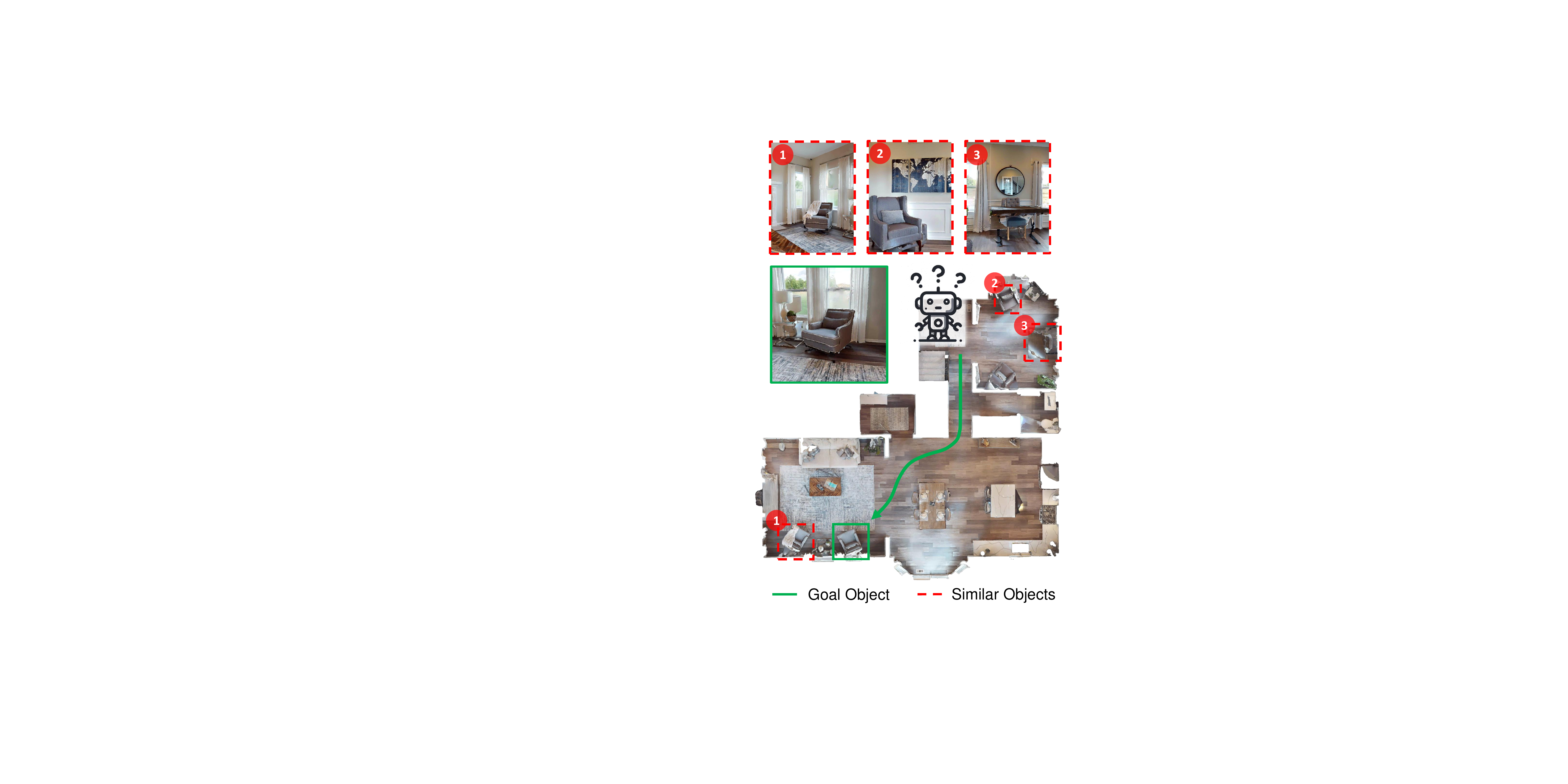}
  \caption{Illustration of Instance ImageGoal Navigation (IIN), which requires agent to navigate to the object instance depicted in the goal image, while distinguishing it from other visually similar instances. 
  }
  \vspace{-16pt}
  \label{fig:intro}
\end{figure}

\IEEEPARstart{E}{mbodied} visual navigation is an emerging computer vision problem where an agent uses visual sensing to actively interact with the world and perform navigation tasks \cite{cheng2022learning, krantz2022sim, zhang2022generative, lin2022multimodal, qiao2023hop+, cai2021pose, lin2021adversarial, an2024etpnav, wu2023human, wang2020vision, wang2023learning, wang2022towards}. Recent years have witnessed substantial progress in embodied visual navigation, fueled by the availability of large-scale photo-realistic 3D scene datasets \cite{xia2018gibson, yadav2023habitat, chang2017matterport3d} and fast simulators for embodied navigation \cite{szot2022habitat, kolve2017ai2, savva2019habitat}. These advancements have enabled researchers to develop and test navigation algorithms in controlled environments that closely mimic real-world conditions. 

In embodied visual navigation, one critical question is ``how can we describe the goal target''.  One common setting for tasking an agent is to give it natural language instructions, such as ``check if my laptop is on the chair". However, this setting becomes confusing when there are multiple chairs in the house. To overcome this challenge, Krantz~\etal \cite{krantz2022instancespecific} propose Instance ImageGoal Navigation (IIN) \cite{krantz2023navigating, bono2023endtoend, lei2024instanceaware}. In IIN, an agent is presented with an image of a specific object, and its goal is to navigate to the specific object within the least time budget. The goal image is not expected to match the sensor specification or embodiment of the navigating agent, as described in \Cref{fig:intro}. To accomplish the task, the agent needs to distinguish the target object from different angle of views and ignore potential distractors. This is a challenging task as it involves semantic reasoning, geometry understanding and instance-aware matching.

To address the above issue, previous methods \cite{chaplot2020learning, chaplot2020object, krantz2023navigating, ramakrishnan2022poni, al2022zero, ma2022end, lei2024instanceaware} introduce 2D Semantic Bird-Eye-View (BEV) map to tackle this problem. These well-designed explicit map representations are memory-efficient, storing essential information such as 2D geometry and semantics, and can be directly utilized to calculate the agent's subsequent actions. However, this simple 2D BEV map representation lacks the capacity to retain 3D geometrical information about the environment, rendering it ineffective for navigating cross-floor scenarios. Additionally, BEV maps are unable to preserve the instance-aware features in a scene, which can be crucial for distinguishing between multiple objects of the same class or for tasks requiring fine-grained object interaction. 


To avoid the limitation of BEV maps, we propose a new Gaussian Splatting Navigation framework, \textit{i.e.}, GaussNav, for IIN task. Our GaussNav is inspired by the recent advancements in 3D Vision technologies, including Neural Radiance Fields (NeRF) \cite{mildenhall2020nerf} and 3D Gaussian Splatting (3DGS) \cite{kerbl3Dgaussians}. These technologies have demonstrated substantial progress in novel view synthesis (NVS) and 3D scene understanding. Although these technologies were not initially developed for navigation tasks, they offer considerable potential for application in this domain. To this end, we develop the Semantic Gaussian map representation, which integrates the representation of geometry, semantics and instance-aware features, and can be directly used for visual navigation. 

Our GaussNav framework consists of three stages, including Frontier Exploration, Semantic Gaussian Construction and Gaussian Navigation. 
First, the agent employs Frontier Exploration to collect observations of the unknown environment.
Second, the collected observations are used to construct Semantic Gaussian. By leveraging semantic segmentation algorithms \cite{he2017mask, wang2023internimage}, we assign semantic labels to each Gaussian. We then cluster Gaussians with their semantic labels and 3D positions, segmenting objects in the scene into different instances under various semantic categories. This representation is capable of preserving not only the 3D geometry of the scene and the semantic labels of each Gaussian, but also the texture details of the scene, thereby enabling NVS. 
Third, we render descriptive images for object instances, matching them with the goal image to effectively locate the target object. Upon determining the predicted goal object's position, we can efficiently transform our Semantic Gaussian into grid map and employ path planning algorithms to accomplish the navigation. 

To the best of our knowledge, we are the first to introduce 3DGS \cite{kerbl3Dgaussians} to embodied visual navigation. In this work, we unify the map representation of geometry, semantics and instance-aware features for visual navigation. Our framework can directly ground the target object with a single goal image input and guide the agent towards it without any additional exploration or verification \cite{lei2024instanceaware}. Our framework designs are beneficial for effective and efficient visual navigation. We evaluate our method's on both efficacy and efficiency and establish new state-of-the-art records on the challenging Habitat-Matterport 3D dataset (HM3D) \cite{yadav2023habitat}.

%% file: sections/related_work.tex
\section{Related Work}

We briefly discuss related work on differentiable rendering, followed by a broad overview of embodied visual navigation, and finally, we focus on the work most relevant to us: Instance ImageGoal Navigation (IIN).

\textbf{Differentiable Rendering.} To achieve photo-realistic scene capture, differentiable volumetric rendering has gained prominence with the introduction of Neural Radiance Fields (NeRF) \cite{mildenhall2020nerf}. NeRF utilizes a single Multi-Layer Perceptron (MLP) to represent a scene, performing volume rendering by marching along pixel rays and querying the MLP for opacity and color. Due to the inherent differentiability of volume rendering, the MLP representation is optimized to minimize rendering loss using multi-view information, resulting in high-quality novel view synthesis (NVS). The primary limitation of NeRF is its slow training speed. Recent advancements have addressed this issue by incorporating explicit volume structures, such as multi-resolution voxel grids \cite{fridovich2022plenoxels, liu2020neural, sun2022direct} and hash functions \cite{muller2022instant}, to enhance performance.

In contrast to NeRF, 3DGS \cite{kerbl3Dgaussians} employs differentiable rasterization. Unlike ray marching, which iterates along pixel rays, 3DGS iterates over the primitives to be rasterized, similar to conventional graphics rasterization. By leveraging the natural sparsity of a 3D scene, 3DGS achieves an expressive representation capable of capturing high-fidelity 3D scenes while offering significantly faster rendering. 
Comprehensive review of the developments in 3D Gaussian Splatting \cite{chen2024survey} highlights the method's versatility and applications across various domains.
Leveraging these advantages, a growing body of research begins to explore various innovations, including deformable or dynamic Gaussians \cite{luiten2023dynamic, wu20234dgaussians, yang2023deformable3dgs}, advancements in mesh extraction and physics simulation \cite{feng2024gaussian,xie2023physgaussian,guedon2023sugar}, as well as applications in Simultaneous Localization and Mapping (SLAM) \cite{keetha2023splatam, matsuki2023gaussian, yugay2023gaussianslam}. This surge of interest in the capabilities of 3DGS leads us to consider its potential for embodied visual navigation. Given that the Gaussian representation inherently encapsulates explicit scene geometry and the parameters required for rendering, we posit that 3DGS can enhance decision-making in map-based visual navigation. The explicit nature of the Gaussian representation provides a rich, condensed form of environmental data, which can be effectively utilized to inform and guide autonomous agents in navigation.

\textbf{Embodied Visual Navigation.} Embodied Visual navigation includes several topics: ObjectGoal Navigation (ObjectNav), Multi ObjectGoal Navigation (MultiON), ImageGoal Navigaiton (ImageNav), and Instance ImageGoal Navigation (IIN). ObjectNav \cite{majumdar2022zson, chaplot2020object, wang2024find, chang2020semantic, chaplot2021seal} requires an agent to navigate to any instance of a specified object category within the environment. MultiON \cite{chen2022learning, wani2020multion, schmalstieg2022learning, marza2023multi}, on the other hand, tasks the agent to sequentially navigate to a series of objects. ImageNav \cite{al2022zero, choi2021image, yadav2023ovrl, yadav2023offline, Kwon_2023_CVPR,Chaplot_2020_CVPR,savinov2018semi,TSGM, wasserman2022lastmile} involves navigating to the camera pose from which a target image is captured. In contrast, IIN \cite{krantz2023navigating, bono2023endtoend, lei2024instanceaware} requires navigating to the specific instance captured by the camera in the target image. Collectively, these navigation tasks span a spectrum from semantic-level navigation in ObjectNav and MultiON to fine-grained instance-level navigation in ImageNav and IIN, comprehensively capturing the problem space of embodied visual navigation. 

Numerous approaches to solving embodied visual navigation utilize deep reinforcement learning (DRL) to develop end-to-end policies that map egocentric vision to action \cite{al2022zero, majumdar2022zson, yadav2023offline, zhu2017target}. However, acquiring skills related to visual scene understanding, semantic exploration, and long-term memory are challenging in an end-to-end framework. Consequently, these methods often incorporate a combination of careful reward shaping \cite{choi2021image}, pre-training routines \cite{yadav2023offline}, and advanced memory modules \cite{savinov2018semi, mezghan2022memory, wu2019bayesian}. In contrast to end-to-end DRL, alternative approaches decompose the problem into sub-tasks that can be optimized in a supervised manner. These sub-tasks include graph prediction via topological SLAM \cite{Chaplot_2020_CVPR}, graph-based distance learning \cite{hahn2021no, shah2021rapid}, and camera pose estimation for last-mile navigation \cite{wasserman2022lastmile}. Chaplot \etal \cite{chaplot2020object} decompose the embodied visual navigation task into exploration, object detection, and local navigation. Building on this, CLIP on Wheels (CoW) \cite{gadre2023cows} utilizes a similar decomposition strategy, focusing on exploration and object localization to handle an open-set object vocabulary. Modular approaches also show promise for effective simulation-to-reality transfer (Sim2Real). Gervet \etal \cite{gervet2023navigating} conduct a Sim2Real transfer of both modular and end-to-end systems, demonstrating that modularity effectively mitigates the visual Sim2Real gap that impairs the performance of end-to-end policies. 

Modular approaches typically represent the environment using a map and utilize this map to acquire the knowledge necessary for navigation within that environment. Bird's-Eye View (BEV) maps~\cite{liu2023bird, chaplot2020object, krantz2023navigating, lei2024instanceaware} are a form of metric map that project the entire scene from an overhead perspective, representing the occupancy status, exploration state, and semantic categories of specific areas. This map representation encodes information about each region into a grid, achieving accurate spatial awareness. In contrast, topological maps~\cite{Chaplot_2020_CVPR, TSGM, Kwon_2023_CVPR} focus more on describing the spatial relationships between different areas in a scene. This approach allows agents to navigate based on the connectivity of spaces rather than relying solely on geometric coordinates. To represent a scene more meticulously, 3D-aware maps have been proposed~\cite{chaplot2021seal, zhang20233d, tan2024self, liu2024volumetric}. Incorporating 3D consistency can enhance perception in navigation. Building upon 3D maps, we further advance by adopting 3D Gaussian Splatting (3DGS)~\cite{kerbl3Dgaussians} as a novel map representation. This enables the map to synthesize appearance views of specific object instances, which we have demonstrated to be effective in the Instance ImageGoal Navigation task.

\textbf{Instance ImageGoal Navigation.} The IIN task introduces distinct challenges compared to the ImageGoal Navigation. First, in IIN, the goal image must depict a specific object instance. In contrast, ImageGoal Navigation may use randomly captured photos that could include insignificant elements, such as large white walls. Second, the camera parameters used to capture the goal image do not necessarily match those of the agent's camera. Therefore, to succeed in IIN, an agent must be adept at identifying the target object among numerous candidates of the same class with goal object and recognizing it from various viewpoints. To address the above challenges, Krantz \etal \cite{krantz2023navigating} develop a general pipeline for aligning the same object from different angle of views. Bono \etal \cite{bono2023endtoend} present an end-to-end approach in the IIN task, while Lei \etal \cite{lei2024instanceaware} propose a method that mimics human behaviour for verifying objects at a distance. Existing methods focus partly on designing sophisticated modules \cite{krantz2023navigating, lei2024instanceaware} and partly on pre-training on pretext tasks \cite{bono2023endtoend}. Our approach differs in that we concentrate on designing a new map representation. Through this novel form of map, we can better establish the connection between target description and target locations, thereby facilitating visual navigation.

%% file: sections/methods.tex
\section{Methods}

\begin{figure}[tb]
\centering
\includegraphics[width=0.95\linewidth]{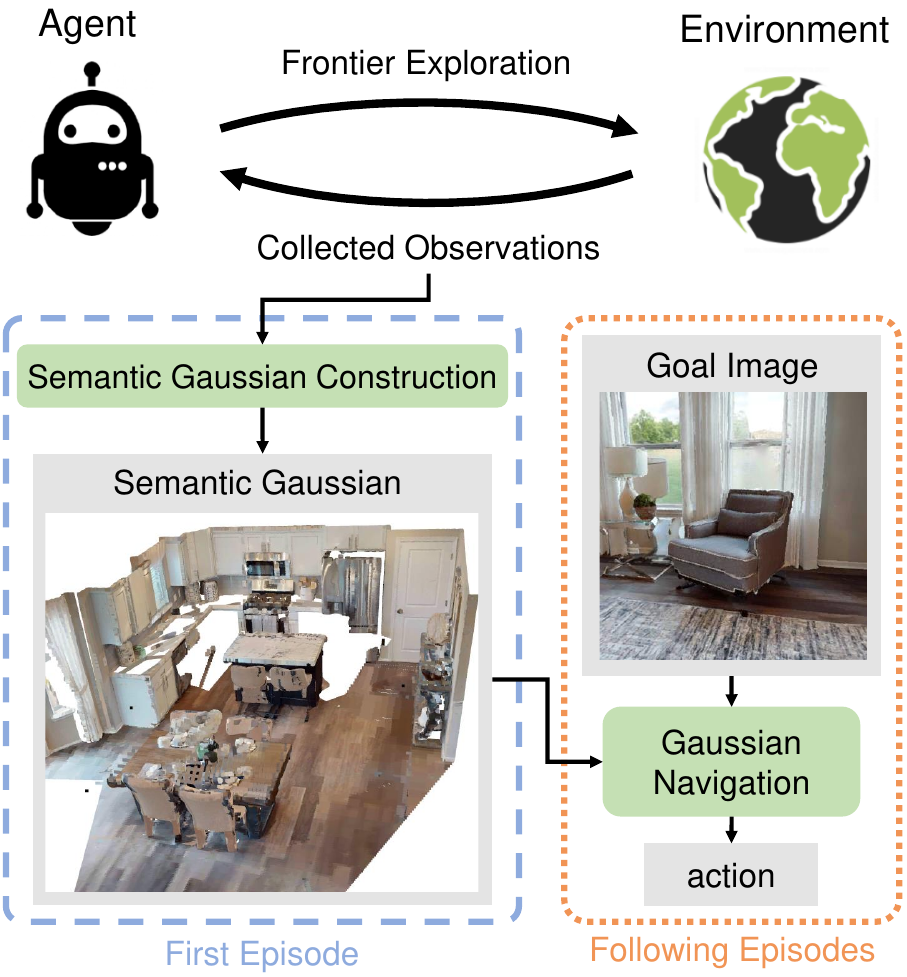}
\caption{Framework Overview. In the first episode of a scene, the agent uses frontier exploration to gather observations of the unknown environment, constructing a Semantic Gaussian. In subsequent episodes, the pre-constructed Semantic Gaussian is utilized by Gaussian Navigation to ground the goal object and guide the agent towards it. }
\vspace{-6pt}
\label{fig:framework_overview}
\end{figure}

\begin{figure}[tb]
\centering
\includegraphics[width=0.98\linewidth]{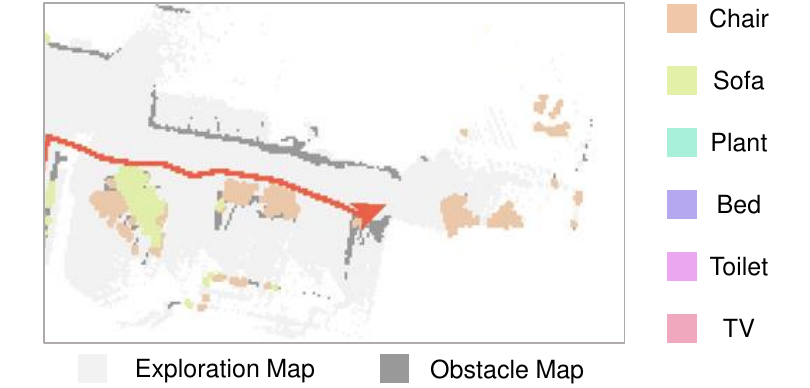}
\caption{Exploration Map and Obstacle Map.}
\vspace{-16pt}
\label{fig:exp_obs_map}
\end{figure}


\subsection{Overview}

In the IIN task, at the start of a new episode $e$, the agent is given a goal image $I_g$ that features a specified object instance $O_g$. The agent's goal is to navigate to the referred instance $O_g$ within the environment. At each timestep $t$, the agent acquires observations which include an RGB image $I_t$, a depth image $D_t$, and sensor pose reading $P_t$. Utilizing this information, the agent must decide upon and execute an action $a_t$. When calling stop action, the episode is considered successful only if the agent is within a certain range of the goal object.

To accomplish the IIN task, we propose a modular framework called Gaussian Splatting for Visual Navigation (GaussNav), as depicted in \Cref{fig:framework_overview}. In a new environment, the Instance ImageGoal Navitation in our proposed framework consists of three stages: Frontier Exploration, Semantic Gaussian Construction, as depicted in \Cref{fig:gauss_construct}, and Gaussian Navigation, illustrated in \Cref{fig:main}. Initially, during the first episode within an unknown environment, the agent employs frontier exploration to explore the environment and collect observations. We then use our proposed Semantic Gaussian to reconstruct the scene. Subsequently, in the following episodes, the agent leverages the Semantic Gaussian to ground the object instance $o$ depicted in the goal image and navigate to it. This process effectively transforms the IIN task into a more manageable PointGoal Navigation task.

\subsection{Frontier Exploration}

In the first episode of an unexplored environment, the agent simultaneously maintains two types of maps, an exploration map and an obstacle map, as illustrated in \Cref{fig:exp_obs_map}. The exploration map delineates the regions of the environment that have been explored, while the obstacle map marks the obstacles in the scene. By detecting the contours of the exploration map and excluding areas in the obstacle map, the agent sets the closest frontier point as a waypoint to facilitate exploration. The frontier-based exploration strategy is a well-established approach in robotics and autonomous navigation \cite{yamauchi1997frontier, holz2010evaluating}. It involves identifying the boundaries between the explored and unexplored regions of the environment, known as frontiers. The agent then selects the nearest accessible frontier point as its next target for exploration. This decision is based on the distance to the frontier, the navigability of the path, and the potential information gain from exploring that frontier \cite{julia2012comparison}. By iteratively exploring the nearest frontiers, the agent efficiently covers the entire environment while avoiding obstacles and previously explored areas. Through the application of this frontier-based exploration strategy, the agent collects observations of the entire environment.

\subsection{Semantic Gaussian Construction}

\begin{figure}[tb]
\centering
\includegraphics[width=0.8\linewidth]{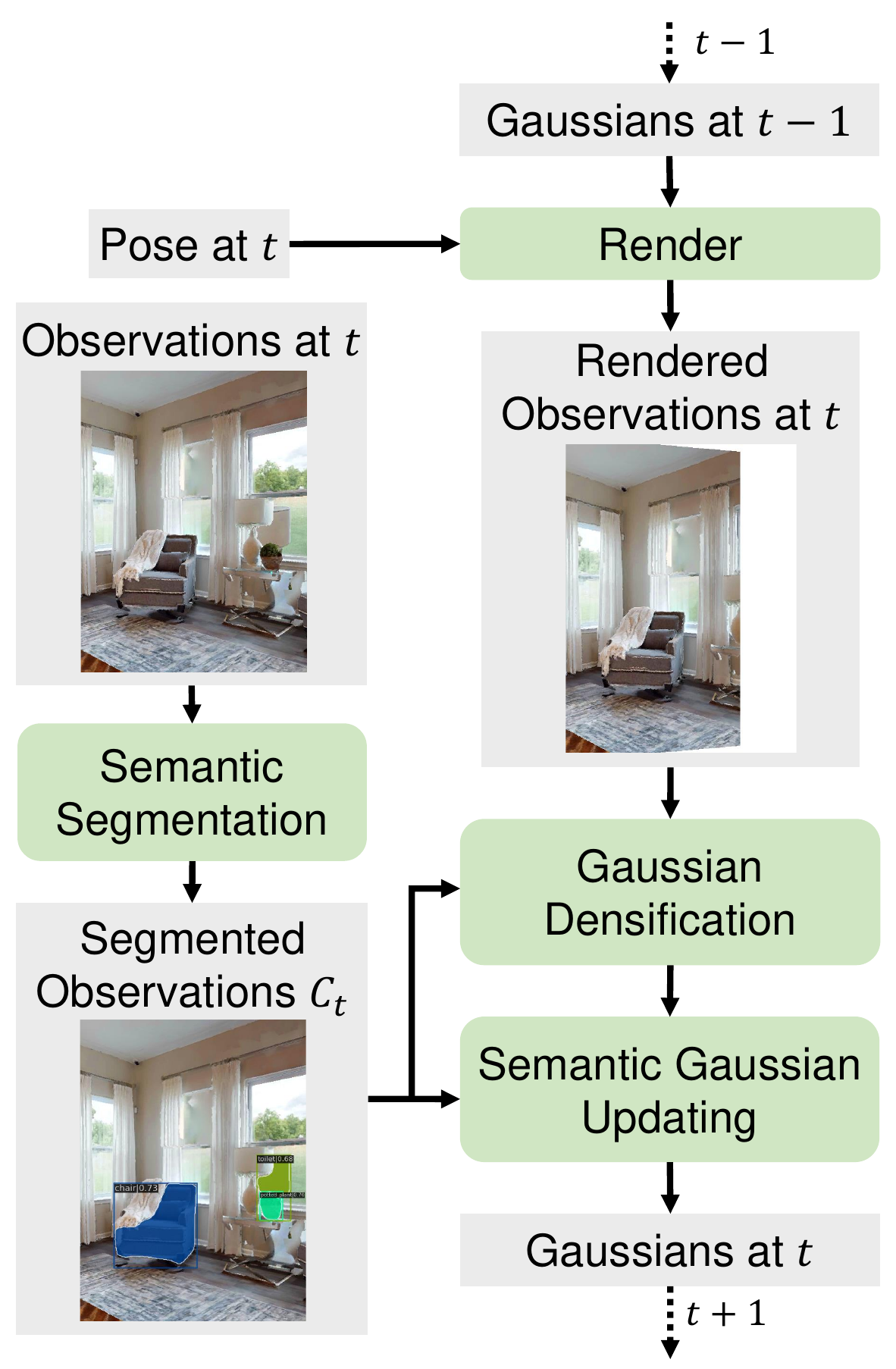}
\caption{An illustration of Semantic Gaussian Construction. At timestep $t$, the pipeline updates the Gaussians from $t-1$ through densification and updating, which involves a comparison between the rendered RGB and depth images against the current input training views. Concurrently, semantic labels are assigned to the densified Gaussians using the segmented images. Finally, the Gaussians are refined through differentiable rendering.}
\label{fig:gauss_construct}
\vspace{-10pt}
\end{figure}

Our Semantic Gaussian is represented by a group of Gaussians, with each Gaussian characterized by a minimal set of nine parameters: a triplet for the RGB color vector $\mathbf{c}$, a triplet delineating the centroid $\bm{\mu} \in \mathbb{R}^3$, a scalar representing the radius $r$, a scalar quantifying the opacity $o \in [0,1]$, and a scalar representing the category label $l$. Different from original 3DGS \cite{kerbl3Dgaussians}, we simplify the representation of Gaussian by using only view-independent color and constraining the Gaussian to be isotropic. This simplification enhances computational efficiency while reducing memory requirements. For a comprehensive understanding of 3DGS \cite{kerbl3Dgaussians}, we highly recommend readers consulting the original paper \cite{kerbl3Dgaussians}. Our Semantic Gaussian Construction can be described as an iterative process comprising two alternating steps: Gaussian Densification and Semantic Gaussian Updating, as depicted in \Cref{fig:gauss_construct}. Gaussian Densification initializes new Gaussians in the Semantic Gaussian at each new incoming frame, while Semantic Gaussian Updating refines the parameters of each Gaussian through Differentiable Rendering.

\textit{Differentiable Rendering.} 3DGS \cite{kerbl3Dgaussians} renders an RGB image as follows: given a collection of 3D Gaussians and camera pose, first sort all Gaussians from front to back. RGB images can then be efficiently rendered by alpha-compositing the splatted 2D projection of each Gaussian in order in pixel space. The rendered color of pixel ${\bf p} = (u,v)$ can be written as:
\begin{equation}
\hat{I}(\mathbf{p}) = \sum_{i=1}^{n} \mathbf{c}_i f_i(\mathbf{p}) \prod_{j=1}^{i-1} (1 - f_j(\mathbf{p})),
\end{equation}
where $f_i(\bf{p})$ is computed as follows:
\begin{align}
    f(\mathbf{x}) = o \exp\left(-\frac{\|\mathbf{x} - \boldsymbol{\mu}\|^2}{2r^2}\right).\label{eq:gauss}
\end{align}
The $\boldsymbol{\mu}$ and $r$ are the splatted 2D Gaussians in pixel-space: 
\begin{equation}
    \boldsymbol{\mu}^{\textrm{2D}} = K \frac{E_t \boldsymbol{\mu}}{d}, \;\;\;\;\;\;
r^{\textrm{2D}} = \frac{f r}{d}, %
\quad \text{where} \quad d = (E_t \boldsymbol{\mu})_z.
\end{equation}
Here, $K$ is the camera's intrinsic matrix, $E_t$ embodies the extrinsic matrix that encodes the camera's rotation and translation at time $t$,  $f$ denotes the known focal length, and $d$ is the depth of the $i^{\text{th}}$ Gaussian relative to the camera's coordinate frame. 

\textbf{Render.} Different from 3DGS \cite{kerbl3Dgaussians}, we differentiably render depth and silhouette image, which determines the visibility and will be used for the next Gaussian Densification and Updating. The depth $D$ and silhouette image $S$ at pixel $\mathbf{p}$ is rendered as follows:
\begin{equation}
    \hat{D}(\mathbf{p}) = \sum_{i=1}^{n} d_i f_i(\mathbf{p}) \prod_{j=1}^{i-1} (1 - f_j(\mathbf{p})), 
\end{equation}
\begin{equation}
    \hat{S}(\mathbf{p}) = \sum_{i=1}^{n} f_i(\mathbf{p}) \prod_{j=1}^{i-1} (1 - f_j(\mathbf{p})).
\end{equation}
Above these, we also render the semantic segmentation results as follows:
\begin{equation}
    \hat{C}(\mathbf{p}) = \sum_{i=1}^{n} l_i f_i(\mathbf{p}) \prod_{j=1}^{i-1} (1 - f_j(\mathbf{p})).
\end{equation}

\textbf{Semantic Segmentation.} As depicted in \Cref{fig:gauss_construct}, the agent's RGB observation $I_t$ is segmented into $C_t$ using Mask-RCNN \cite{he2017mask}. The segmented result will be used for initializing new Gaussians in Gaussian Densification and supervising the Gaussians' parameters in Semantic Gaussian Updating.

\textbf{Gaussian Densification.} Gaussian Densification is performed by comparing the rendered results at position $P_t$ using Gaussians from $t-1$ with the ground truth. This process adds new Gaussians where the previous Gaussians fail to represent the scene in the new observations. Following Keetha \etal \cite{keetha2023splatam}, we add new Gaussians based on a densification mask to determine which pixels should be densified:
\begin{equation}
\small
M(\textbf{p}) = \Bigl(\hat{S}(\textbf{p}) < 0.5\Bigr) + \Bigl( D(\textbf{p}) < \hat{D}(\textbf{p}) \Bigr) \Bigl(\textrm{L}_1\bigl(\hat{D}(\textbf{p})\bigr) > 50 \textrm{MDE} \Bigr),
\end{equation}
where the first term represents where the Semantic Gaussian is not adequately dense, and the second term indicates where the ground-truth depth is in front of the predicted depth and the depth error is greater than 50 times the median depth error (MDE).

\textbf{Semantic Gaussian Updating.}  After densifying current Semantic Gaussian, we update the parameters of Gaussians given poses and observations. This is done by differentiable-rendering and gradient-based-optimization, which is equivalent to the ``classic'' problem of fitting a radiance field to images with known poses. Specifically, we update the parameters of Gaussians by minimizing the RGB, depth and segmentation errors.

\begin{figure}[tb]
\centering
\includegraphics[width=0.9\linewidth]{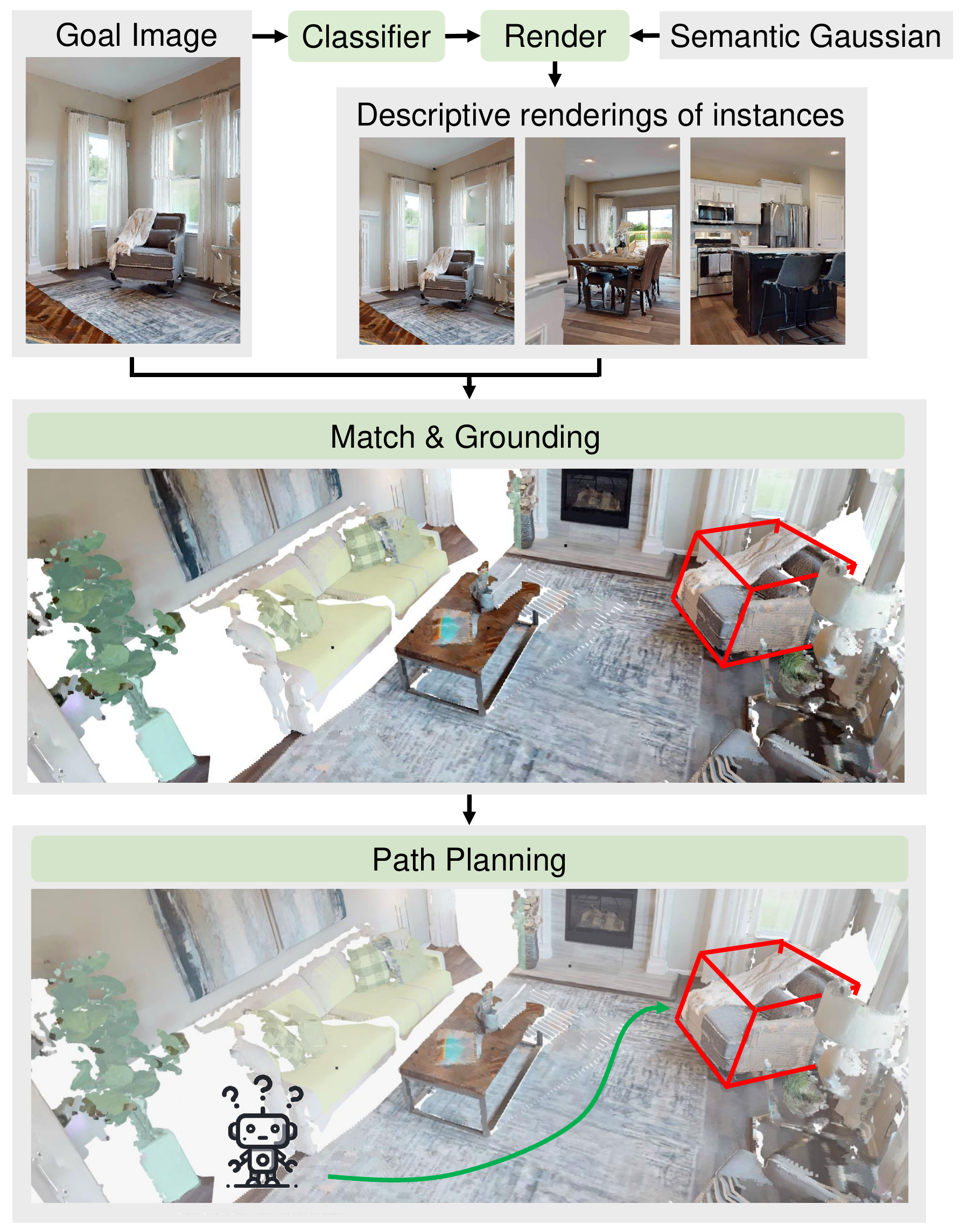}
\caption{An illustration of Gaussian Navigation. Our approach begins with the classification of the goal image using pre-constructed Semantic Gaussian. Upon determining the predicted class, we generate descriptive images around instances belonging to that class. These images are then matched with the target object to identify and ground the goal instance. Utilizing the map and the established goal, the agent employs path planning to compute the sequence of actions.}
\label{fig:main}
\vspace{-10pt}
\end{figure}

\subsection{Gaussian Navigation}

To navigate using a constructed Semantic Gaussian, we propose Gaussian Navigation, as illustrated in \Cref{fig:main}. We first classify the goal image $I_g$ to predict the semantic label $\hat{l}_g$ of the current target, such as `chair' in \Cref{fig:main}. This semantic label is then used to query the relevant Gaussians. For each relevant object instance of the same class, we render descriptive images. These renderings are compared with the goal image to ground the goal object, predicting its position $\hat{P}_g$. Finally, the Path Planning module generates a feasible path and determines the agent's action.


\textbf{Classifier.} In IIN task, the agent receives an image depicting the target object $I_g$. However, comparing this image with renderings from navigable points within the entire scene becomes exceedingly time-inefficient due to the vastness of the search space. Consequently, with our Semantic Gaussian $G_s$, we only need to search for instances of the object category corresponding to the goal object. Therefore, we first classify the goal image $I_g$ into target category label $\hat{l}_g$. We use the goal images on the train split of HM3D-SEM \cite{yadav2023habitat} to finetune the image classification model, \textit{i.e.}, ResNet50 \cite{He2015}, pretrained on ImageNet \cite{5206848}. 


\textbf{Match \& Grounding.} With the predicted target object's label $\hat{l}_g$, we identify all candidate objects that share the same class label. For each candidate instance, we generate a set of descriptive images by rendering the object from multiple viewpoints to capture its features. Specifically, for one training view containing possible candidate objects, we augment it to $n_v$ views by novel view synthesis (NVS). we denote the transformation from the camera to the world coordinate system of the training view as $\text{c2w}$, and record the translation of the potential target object in the camera frame as $\mathbf{t}^{o}_{c}$. We define the rotation matrix from the object to the world frame using the \textit{forward}, \textit{right}, and \textit{up} vectors. The \textit{forward} direction is the vector from the training view to the potential target object, namely $\mathbf{t}^{o}_{c}$. The \textit{up} direction is the vector $[0,\,-1,\,0]$, and the \textit{right} vector is orthogonal to both the \textit{forward} and \textit{up} vectors, forming a right-handed coordinate system with the \textit{right}, \textit{up}, and \textit{forward} vectors. The translation vector from the object to the world frame is
    \begin{equation}
        \mathbf{t}^{o}_{w} = \text{c2w} \times \mathbf{t}^{o}_{c}.
        \label{equ:NVS0}
\end{equation}
Together, the rotation matrix and translation vector constitute the rigid transformation matrix $\text{o2w}$ from the object to the world frame.

We define the rotation matrices around the $y$-axis and $x$-axis as $\mathbf{R}_y(\theta)$ and $\mathbf{R}_x(\theta)$, respectively, representing new viewpoints formed by rotating around the object by an angle $\theta$ in the horizontal and vertical directions. Thus, the transformation from the camera to the world frame for new viewpoints in the horizontal direction is
\begin{equation}
    \text{c2w}_h(\theta) = \text{o2w} \times \mathbf{R}_y(\theta) \times \text{w2o} \times \text{c2w},
    \label{equ:NVS1}
\end{equation}
and in the vertical direction:
\begin{equation}
    \text{c2w}_v(\theta) = \text{o2w} \times \mathbf{R}_x(\theta) \times \text{w2o} \times \text{c2w}.
    \label{equ:NVS2}
\end{equation}
In experiments, when $n_v=1$, we do not perform NVS; when $n_v=3$, we perform NVS at $\theta = \pm15^\circ$ (both horizontal and vertical); and when $n_v=5$, we use $\theta = \pm15^\circ$, $\pm30^\circ$.

After NVS, the original training views are augmented. The augmented rendering set for the $i$-th object instance is denoted as $S_i$. Let $n$ denote the observed instances of the same class as target object, then the universal set of $S$ can be formulated as:
\begin{equation}
    S = \{S_1, S_2, \ldots, S_n\}  \quad \text{for} \quad i=1,2,\ldots,n.
\end{equation}
To distinguish the target object from these candidates, we can formulate the question as:
\begin{equation}
\begin{split}
      i_{\text{max}} = arg\max \{ \max_{s \in S_1} \Omega (s), \max_{s \in S_2} \Omega (s), \ldots, \max_{s \in S_n} \Omega (s) \} \\
      \text{for} \quad i=1,2,\ldots,n,
    \label{eq:match}  
\end{split}
\end{equation}
where $\Omega (\cdot)$ is defined as the matched number of keypoints between renderings and goal image $I_g$. Specifically, for the rendering $s \in S_i$ of the $i$-th object instance and the goal image $I_g$, we extract the pixel-wise $(x, y)$ coordinates of the keypoints and their associated feature descriptors $V_t$ using DISK \cite{tyszkiewicz2020disk}. That is:
\begin{equation}
(K_t, V_t) = \text{DISK}(s), \quad (K_g, V_g) = \text{DISK}(I_g).
\label{eq:feature1}
\end{equation}
Subsequently, the matched pairs $(\hat{K_t}, \hat{K_g})$ are computed using LightGlue \cite{lindenberger2023lightglue}. The feature matching process is formulated as follows:
\begin{equation}
(\hat{K_t}, \hat{K_g}) = \text{LightGlue}((K_t, V_t), (K_g, V_g)).
\label{eq:matching}
\end{equation}
Thus, the number of matched points, \textit{i.e.}, the length of $\hat{K_t}$ or $\hat{K_g}$, is denoted as $\Omega$. The candidate object whose rendered images yield the highest number of matched keypoints is then selected as the target object, as shown in \Cref{eq:match}.

When the object instance is selected, we ground the object in the Semantic Gaussian. Due to the presence of outliers, which are caused by errors in semantic segmentation, we perform clustering on the instances on the map. Specifically, we use Density-based Spatial Clustering of Applications with Noise (DBSCAN) \footnote{\url{https://scikit-learn.org/stable/modules/generated/sklearn.cluster.DBSCAN.html}}, which groups together points that are closely packed together, marking as outliers points that lie alone in low-density regions. With the precise selected object instance's location, we can easily transform the IIN task to a PointGoal task.

\textbf{Path Planning.} The Semantic Gaussian is not designed for path planning. We first convert the Semantic Gaussian into a point cloud, whereby each Gaussian is reduced to a single point in the point cloud representation. The point cloud is then voxelized into 3D voxels $M_{3D}$, then the 3D voxels $M_{3D}$ are projected to 2D BEV grids $M_{2D}$. Here, we use the 2D projection of Semantic Gaussian rather than the 2D geometric map used in pre-exploration to maintain a consistent representation throughout the pipeline. Our Semantic Gaussian initializes Gaussians using point cloud derived from depth image and does not prune any Gaussian in the optimization stage. Thus, 2D projection of Semantic Gaussian or 2D geometric map are fundamentally equivalent.

Given 2D BEV grid map $M_{2D}$, along with the agent's starting position and the goal's location, we can efficiently generate a shortest distance field using FMM. Each point within this field encapsulates the minimal distance necessary to traverse from the starting point to the goal.  We extract a relevant subset of this distance field that falls within the agent's operational range. Subsequently, a waypoint is chosen from this subset to ensure it avoids any intersections with obstacles while adhering to a local minimum in the distance field. With the selected waypoint, the agent can readily calculate an action based on the angle and distance from its current state. The agent iterates the aforementioned process to generate a sequence of actions, continuing until the destination is reached.

%% file: sections/experiments.tex
\section{Experiments}

This section first details our experiment setup, followed by a comparative analysis with state-of-the-art approaches. Subsequently, we present an ablation study to evaluate the efficacy of individual components in our proposed methods. Finally, we assess the computational efficiency of our approach, examine error cases, and evaluate the rendering quality.

\input{tables/main_table}

\subsection{Experiment Setup}

\textbf{Datasets.} We use Habitat-Matterport 3D dataset (HM3D) \cite{yadav2023habitat} in the Habitat \cite{szot2022habitat} for our experiments. HM3D consists of scenes which are 3D reconstructions of real-world environments with semantic annotations. These scenes are split into three distinct subsets for training, validation, and testing, consisting of 145/36/35 scenes, respectively. We follow the task setting of Instance ImageGoal Navigation (IIN) proposed by Krantz \etal \cite{krantz2022instancespecific}. The episode dataset has been partitioned into three subsets for training, validation, testing, comprising 7,056K/1K/1K episodes respectively. The object depicted by the goal image belongs to the following six categories:  \{``chair'', ``couch'', ``plant'', ``bed'', ``toilet'', ``television''\}. On the validation subset, a total of 795 unique object instances have been observed.

\textbf{Embodiment.} We adopt the embodiment parameters from the Hello Robot Stretch platform \footnote{\url{https://hello-robot.com/stretch-2}}. The simulated agent is modeled as a rigid-body cylinder with zero turning radius, a height of 1.41m, and a radius of 0.17m. A forward-facing RGB-D camera is affixed at a height of 1.31 m. At each timestep $t$, the agent's observation consists of an egocentric RGB image, depth image, goal image and sensor pose reading. Camera specifications, such as mounting height, look-at angle, and field of view (FOV), differ between the agent's and the goal's cameras. Specifically, the agent's camera resolution is $640\times480$, whereas the goal's camera has a resolution of $512\times512$ with unfixed height and FOV parameters.

\begin{figure}[tb]
\centering
\includegraphics[width=0.45\textwidth]{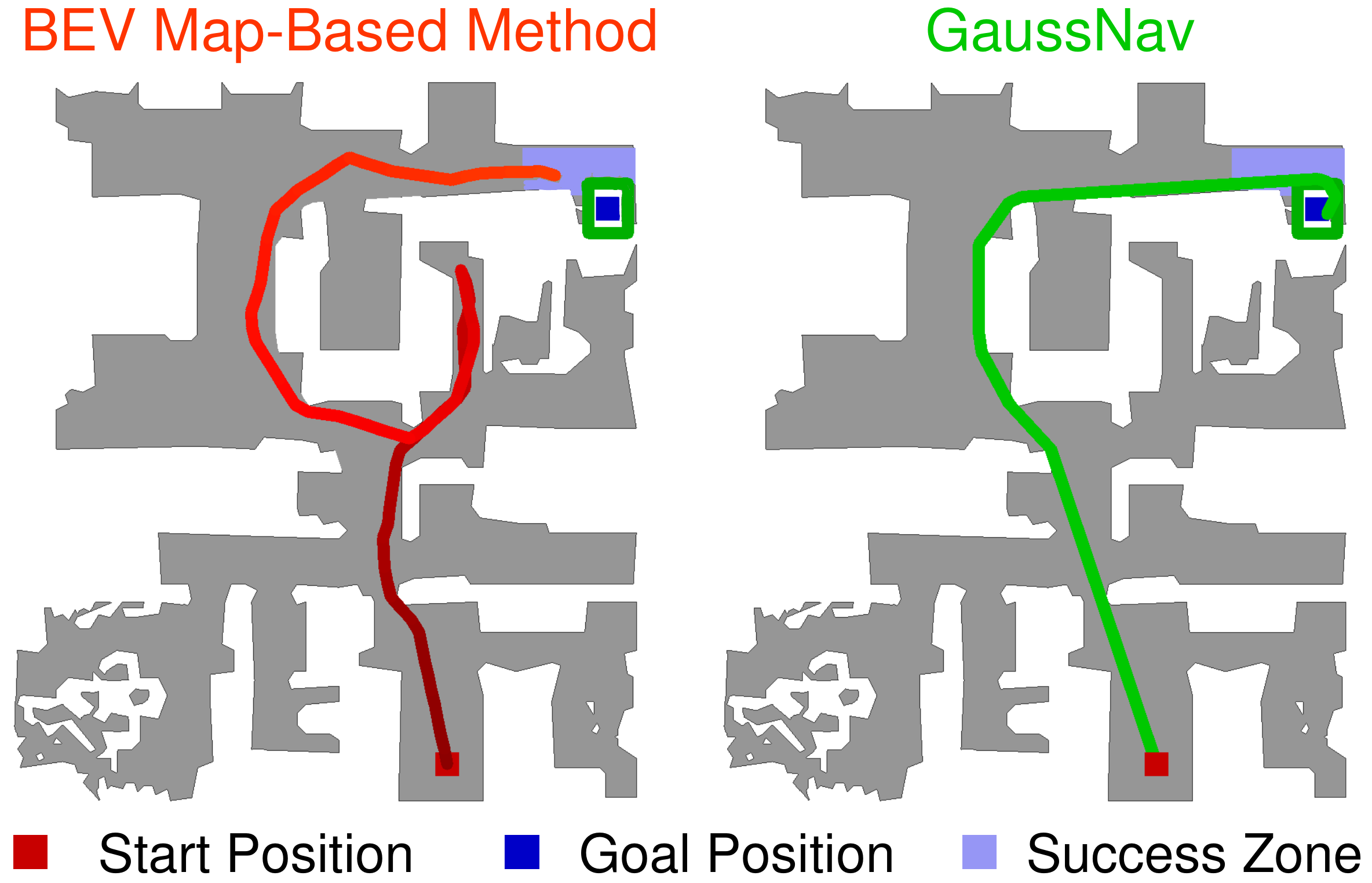}
\caption{Trajectory Analysis. Our Semantic Gaussian map representation can allow agent to directly ground target object from a single goal image, thereby facilitating efficient visual navigation.}
\label{fig:trajectory}
\vspace{-10pt}
\end{figure}

\textbf{Action Space.} We use a discrete action space for navigation, comprising four actions: \{\texttt{STOP}, \texttt{FORWARD}, \texttt{TURN\_RIGHT}, \texttt{TURN\_LEFT}\}. The \texttt{STOP} action terminates the current episode, while the \texttt{FORWARD} action advances the agent by 25 cm. Rotational actions occur in place: \texttt{TURN\_RIGHT} induces a 25-degree clockwise rotation and \texttt{TURN\_LEFT} a 25-degree counter-clockwise rotation.

\textbf{Evaluation Metrics.} Following Krantz \etal \cite{krantz2022instancespecific}, we evaluate our model with both success and efficiency. We report Success Rate (Success), Success rate weighted by normalized inverse Path Length (SPL). An episode is deemed successful ($\text{Success} = 1$) if the agent invokes the \texttt{STOP} action within a $1.0$m Euclidean distance from the goal object. SPL is an efficiency measure defined in \cite{anderson2018evaluation}, is given by:
\begin{equation}
    \text{SPL} = \frac{1}{N} \sum_{i=1}^{N} S_i \cdot \frac{l_i}{\max(p_i, l_i)},
\end{equation}
where $N$ is the total number of episodes, $S_i$ is a binary success indicator for episode $i$, $l_i$ is the shortest path distance from the start position to the goal, and $p_i$ is the path length actually traversed by the agent. A higher SPL value indicates more efficient navigation.

\subsection{Comparison to the State-of-the-art Methods}

We evaluate our proposed model against various baselines and previous state-of-the-art work, as presented in \Cref{tab:main}. Unlike conventional IIN methods that generate a map for each episode, GaussNav constructs a map across episodes in a scene. The first block lists results from end-to-end baselines, and the second block includes methods designed for the MultiON task but implemented here for the IIN task. The third block of \Cref{tab:main} presents the performance of original state-of-the-art IIN methods, while the fourth block reports their performance when adapted to use scene map.

\textbf{End-to-end Baselines.} We evaluate the performance of two end-to-end methods on the IIN task. \textit{RL Baseline} built a network that observes agent RGB $(\mathcal{V}_{RGB})$, agent depth $(\mathcal{V}_{D})$, the goal image $(\mathcal{V}_G)$, GPS coordinates $(x,z)$, and heading $(\theta)$. Visual observations are encoded with separate ResNet-18~\cite{he2016deep} encoders and GPS and heading are encoded with 32-dimensional linear layers. Then, these features are concatenated and encoded with a 2-layer LSTM. Finally, an action $a(t)$ is sampled from a categorical distribution. The network was trained from scratch using Proximal Policy Optimization (PPO) \cite{schulman2017proximal}.

\textit{OVRL-v2}, proposed by Yadav \etal \cite{yadav2023ovrl}, introduced self-supervised pretraining for visual encoders in ImageGoal Navigation. Originally, OVRL-v2 was trained for the ImageGoal Navigation task. Direct application of OVRL-v2 to IIN task without fine-tuning yields suboptimal performance, as indicated by a Success of 0.006 (row 2 in \Cref{tab:main}). This reduced efficacy can be attributed to several factors: the transition between scene datasets from Gibson \cite{xia2018gibson} to HM3D \cite{yadav2023habitat}, differences in robot embodiment from Locobot \footnote{\url{http://www.locobot.org/}} to Stretch, and a shift in the nature of goal destinations from image sources to image subjects. Fine-tuning OVRL-v2 specifically for IIN task on the HM3D dataset significantly improves outcomes, resulting in a Success of 0.248 (row 3 in \Cref{tab:main}).

\textbf{State-of-the-art Methods in MultiON.} The MultiON task shares many similarities with the Scene-specific Map representation we employ in evaluating GaussNav. We re-implement several state-of-the-art methods \cite{wani2020multion, chen2022learning, marza2023multi} from the MultiON task on the IIN benchmark. It is worth noting that we directly input the semantic category of the target object to the aforementioned methods. As the navigation requirements of the IIN task are at the instance level rather than the semantic level, this discrepancy in task formulation leaves room for performance improvement.

\input{tables/ablations}

\input{tables/abla_classifier}

\textbf{State-of-the-art Methods in IIN.} For fair comparison, We evaluate the performance of previous state-of-the-art methods \cite{krantz2023navigating, lei2024instanceaware} on the IIN task using two types of map representations: episodic map and scene-specific map. \textit{Mod-IIN} \cite{krantz2023navigating} decomposes the IIN task into exploration, goal instance re-identification, goal localization, and local navigation. This method utilizes feature matching to re-identify the goal instance within the egocentric vision and projects the matched features onto a map to localize the goal. Each sub-task is addressed using off-the-shelf components that do not require any fine-tuning.

\textit{IEVE} \cite{lei2024instanceaware} employs a modular architecture that dynamically switches between exploration, verification, and exploitation actions. This flexibility empowers the agent to make informed decisions tailored to varying circumstances. \textit{Mod-IIN} and \textit{IEVE} both demonstrate exceptional performance on the IIN task. We also implement the scene-specific map representation in these methods for a fair comparison with GaussNav. As shown in \Cref{tab:main}, GaussNav demonstrates a significant performance advantage over all methods. It significantly surpasses all existing models in terms of SPL by a huge margin of 0.231 (last 2 rows in \Cref{tab:main}). This result indicates that our Semantic Gaussian map, by preserving the intricate texture details of objects in the scene, enables the agent to directly locate the target object based on the goal image without the need for additional verification.

We attribute the superior performance to the novel map representation of Semantic Gaussian, allowing agent to directly ground goal target without additional exploration or verification, as evidenced by \Cref{fig:trajectory}. Unlike previous widely-used BEV map representation, our Semantic Gaussian can allow agent to query Gaussian through semantic label input and render descriptive images of an object instance. Therefore, our GaussNav does not require explicit verification of potential object candidates, unlike BEV map-based approaches. Instead, it selects the most probable candidate from a multitude of possibilities and navigates towards it directly.

\begin{figure}[tb]
\centering
\includegraphics[width=0.45\textwidth]{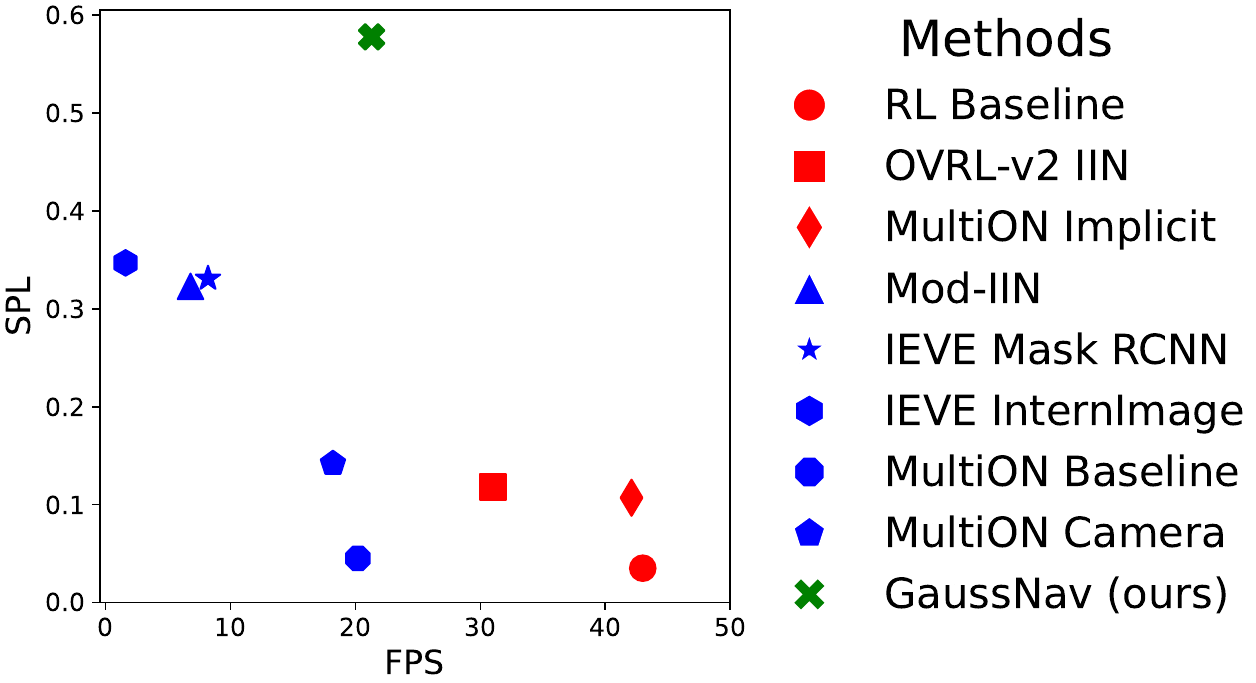}
\caption{SPL and FPS Analysis. {\color{red}Red markers} represent end-to-end methods, while {\color{blue}blue markers} indicate modular approaches. Our {\color{MyGreen}GaussNav} belongs to the modular approach, achieving the highest frame rate among modular methods while attaining the highest SPL across all approaches in IIN task.}
\vspace{-10pt}
\label{fig:framerate}
\end{figure}

\begin{figure*}
  \centering
  \includegraphics[width=0.95\linewidth]{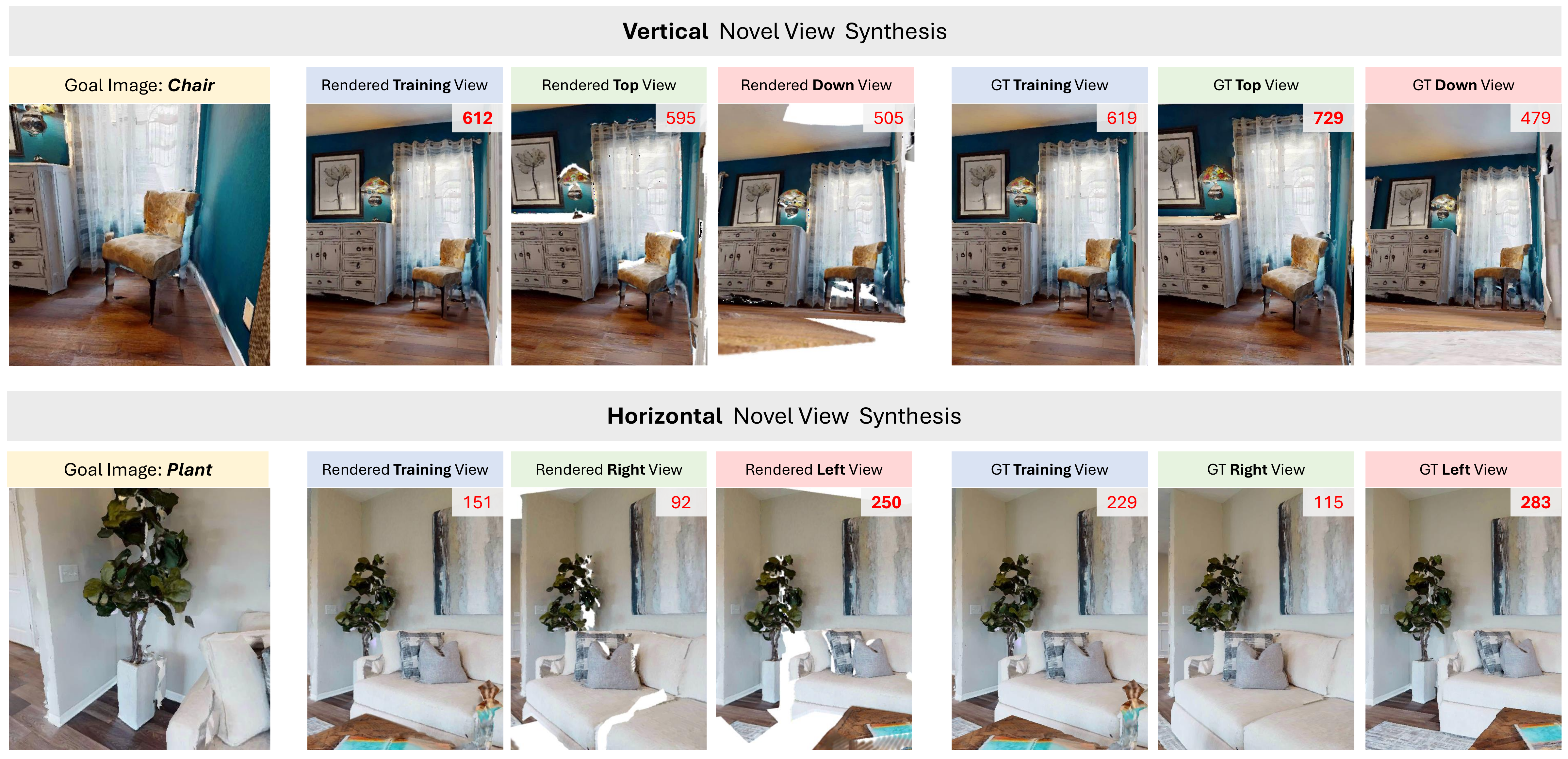}
  \vspace{-10pt}
  \caption{Visualization of novel view  synthesis results. The visualization results include horizontal and vertical novel view  synthesis results with $\theta = \pm15^\circ$. The upper right corner of each image shows the number of matched keypoints with the goal image.
  }
  \vspace{-6pt}
  \label{fig:NVS}
\end{figure*}

\input{tables/time}

\input{tables/NVS}

\subsection{Ablation Study}

To understand the modules of our GaussNav, we consider the following ablations:

\textbf{GaussNav w.o. Classifier.} In \Cref{fig:main}, we replace the Classifier's output with a random generated target category. We observe that the Success drops to 0.375 and the SPL decreases to 0.291 (row 2 in \Cref{tab:ablations}). To better evaluate the classifier, we design the following experiment. We define the training views from the full space of candidate objects as the complete set, and the training views filtered by the classifier as a subset. Given the goal image and the keypoint matcher, we compute, in each set respectively, the view with the largest number of matching keypoints. If this view contains the goal object, we consider it a success; otherwise, a failure. We also record the total time spent in the entire local feature matching process to assess the impact of the classifier on efficiency. Here, we do not use the navigation metrics such as Success and SPL because this allows us to eliminate the influence of irrelevant factors, such as path planning errors. Our experimental results are shown in \Cref{tab:abla_classifier}. As can be seen, the total success without the classifier is slightly higher than that with the classifier (an increase of 0.039), but the improvement is limited. However, the time taken is 2.5 times that with the classifier, making it significantly less efficient. Therefore, considering the trade-off between performance and efficiency, we choose to include the classifier as a component of GaussNav framework.


\textbf{GaussNav w.o. Match.} The Match module is designed to distinguish the target from candidates with the same class. Without the Match module, we randomly select from these candidates. The Success and SPL falls to 0.444 and 0.353 (row 3 in \Cref{tab:ablations}). This can be attributed to the random selection of candidate instances without Match module.

\textbf{GaussNav w. SIFT or GlueStick \cite{pautrat_suarez_2023_gluestick}.} To evaluate the impact of different extractors and matching algorithms on our GaussNav, we replace the combination of DISK \cite{tyszkiewicz2020disk} + LightGlue \cite{lindenberger2023lightglue}. We employ both SIFT + FLANN and GlueStick \cite{pautrat_suarez_2023_gluestick} as alternatives. The first replacement represents a greater decrease, (row 4 in \Cref{tab:ablations}) and the second only displays a slight decrease (row 5 in \Cref{tab:ablations}). These results demonstrate the varying performance of different local feature matching algorithms on the HM3D dataset.

\textbf{NVS Analysis.} We conduct experiments to evaluate the impact of the number of rendered images $n_v$ and the upper bound of NVS. Specifically, we perform ablation studies on NVS with $n_v=3$ and $n_v=5$ in both horizontal and vertical directions, and also evaluate the upper bound achievable using GT rendered results. To avoid the influence of other factors in navigation, we consider it a success only if the image retrieved through keypoint matching contains the target object; otherwise, it is considered a failure. The specific evaluation metrics are detailed in \textbf{GaussNav w.o. Classifier.}. The experimental results are presented in \Cref{tab:NVS}. Overall, utilizing NVS is beneficial for successfully recognizing objects. However, a large $n_v$ does not necessarily yield positive effects (see \Cref{tab:NVS}, vertical $n_v=5$ \textit{vs.} vertical $n_v=3$). This is because our Semantic Gaussian map may have only a few observational viewpoints for a particular object, unlike the dozens available in traditional 3DGS. Therefore, when rendering from novel viewpoints, artifacts such as holes may occur, reducing the number of features, as visualized in \Cref{fig:NVS}. Nevertheless, for GT NVS, the improvement is significant.

\input{tables/ieve_nvs}

To further study the impact of NVS without pre-exploration, we adopt the same framework as IEVE~\cite{lei2024instanceaware}, with the only difference being the map representation. IEVE uses 2D BEV map while GaussNav implements Semantic Gaussian map. Additionally, GaussNav performs NVS whenever a similar object appears in each observation. The results are presented in \Cref{tab:pre-exploration}. Our Semantic Gaussian map enhances the existing observations without relying on pre-exploration. While the improvement may not be as prominent as the ability of instance-level object localization, the result of last row in \Cref{tab:pre-exploration} indicates promising direction for future work.

\subsection{Efficiency Analysis}

\begin{figure*}
  \centering
  \includegraphics[width=0.95\linewidth]{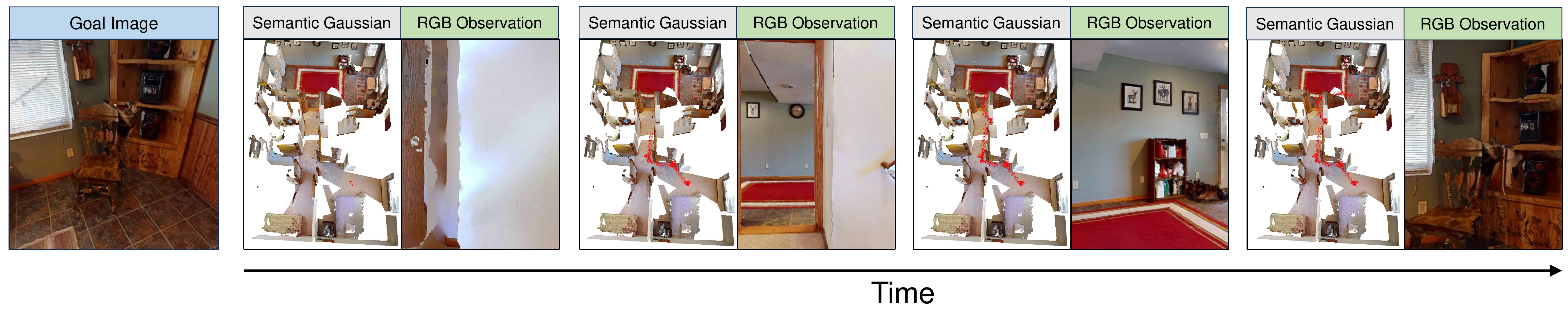}
  \vspace{-10pt}
  \caption{Qualitative example of our GaussNav agent performing IIN task in the Habitat simulator. When randomly initialized in the environment, the agent is given a goal image depicted a target object. Gaussian Navigation directly grounds the target object in the Semantic Gaussian and guide the agent towards it.
  }
  \vspace{-10pt}
  \label{fig:qualitive_result}
\end{figure*}

\begin{figure*}
  \centering
  \includegraphics[width=0.95\linewidth]{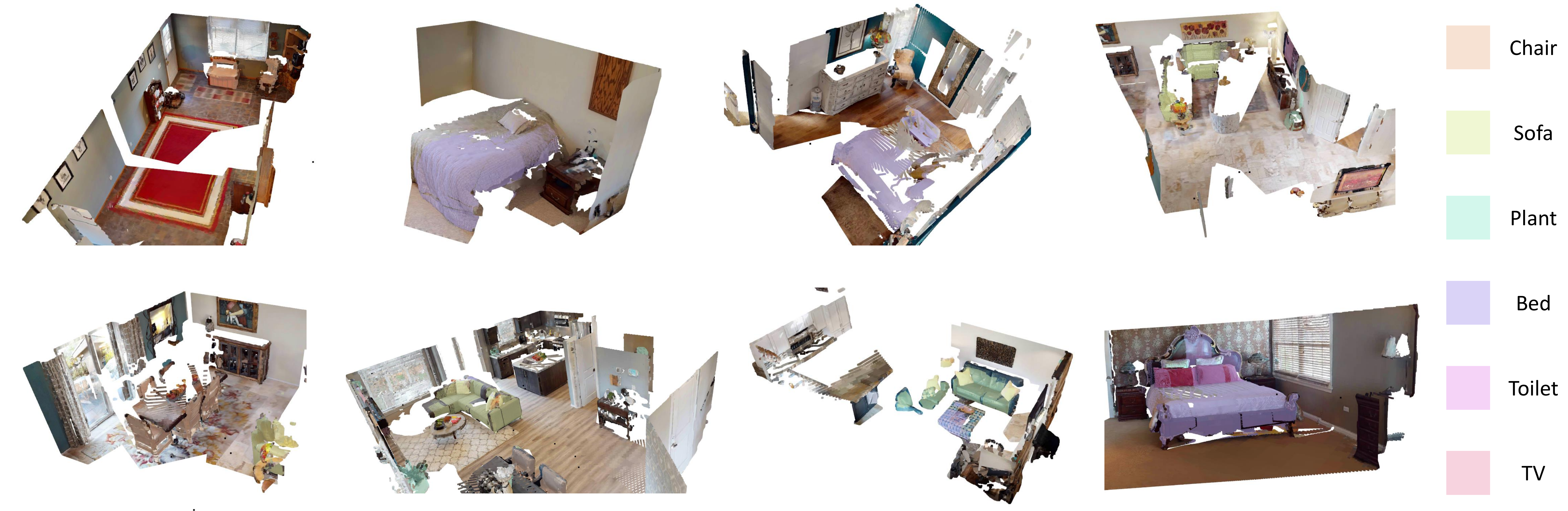}
  \caption{Semantic Gaussian visualization results.
  }
  \vspace{-6pt}
  \label{fig:vis_gauss}
\end{figure*}

We analyze the temporal efficiency of our approach from two perspectives: preprocessing and agent-environment interaction. In the context of our method, ``preprocessing'' refers to the process of matching and grounding target objects given the Semantic Gaussian. The latter perspective compares the runtime frame rate of our method with various other approaches.

Efficiently locating the target object within the map is crucial for our method, which leverages renderings to match and ground the goal object. To reduce the computational cost of comparing all possible observations at each navigable viewpoint, we introduce Semantic Gaussians. This technique groups object instances under their corresponding semantic labels. This optimization significantly reduces the search space by limiting comparisons to several descriptive renderings of each instance. For example, in the scene \texttt{CrMo8WxCyVb}, the navigable area is quantified to be $54.03 m^2$. This space can be discretized into approximately 54 squares, each with an area of $1 m^2$. Positioned within each square, the agent can observe its surroundings from 12 distinct viewing angles, covering a full $360 \deg$ with each angle spanning $30 \deg$. Thus, the original search space for this single floor would consist of $12 \times 54 = 648$ potential observations. By applying our grouping strategy based on semantic labeling, the search space is considerably narrowed. We count the number of different categories of object instances within the \texttt{CrMo8WxCyVb} scene, as demonstrated in \Cref{tab:time}. To locate a ``chair'' — assuming we render each object instance from three unique viewpoints — the resulting search space is reduced to merely $3 \times 11 = 33$ observations. We render three observations of a single instance to ensure rendering quality, which is achievable only when the new viewpoint largely overlaps with the training views. This optimization yields a significant improvement in time efficiency.

\begin{figure*}
  \centering
  \includegraphics[width=0.99\linewidth]{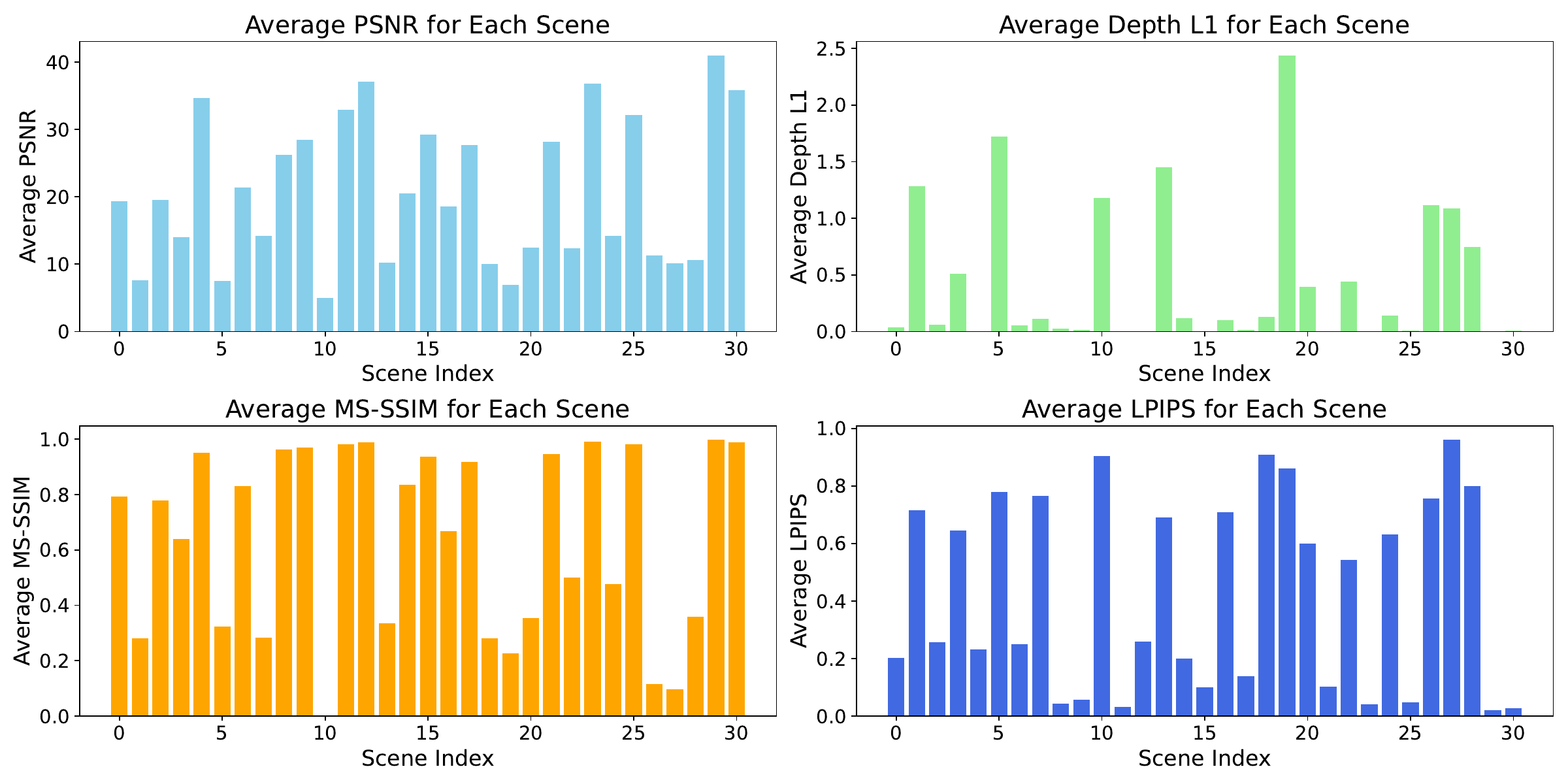}
  \caption{Rendering quality of our Semantic Gaussian Construction results on the HM3D validation dataset. The x-axis indicates different scene indices with the corresponding floor height.
  }
  \vspace{-4pt}
  \label{fig:gauss_results}
\end{figure*}

\begin{figure*}
  \centering
  \includegraphics[width=0.99\linewidth]{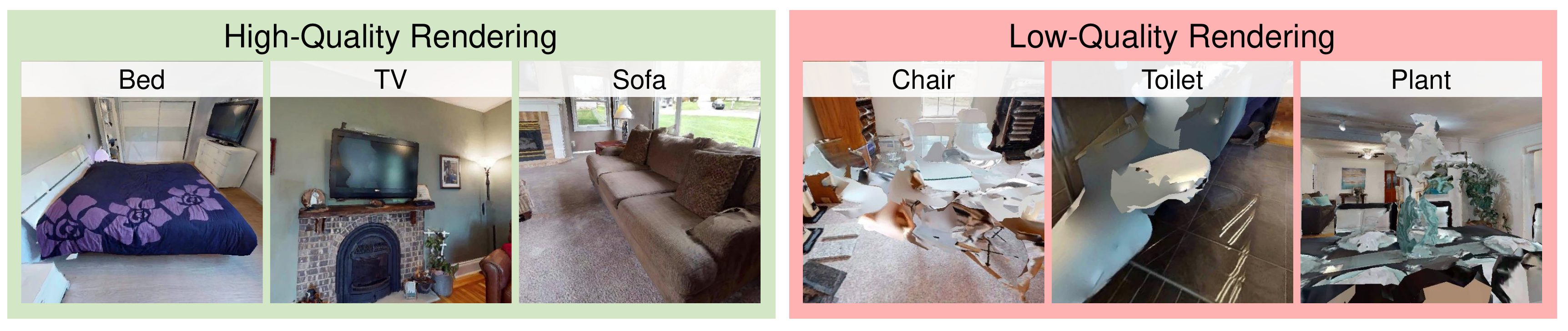}
  \caption{Observations rendered from HM3D \cite{yadav2023habitat} scene dataset using the Habitat \cite{szot2022habitat} simulator.
  }
  \vspace{-16pt}
  \label{fig:bad_render}
\end{figure*}

We also compare the runtime frame rates of different methods in a single Habitat environment using an NVIDIA GeForce RTX 3090 GPU and 10 CPU cores. As shown in \Cref{fig:framerate}, our method maintains a high efficiency of over 20 FPS while achieving the highest SPL among the compared approaches. To achieve higher runtime speed, our GaussNav projects the Semantic Gaussian to obtain a 2D grid map. By utilizing the predicted target location from the ``preprocessing'', we then employ Fast Marching Method (FMM) for path planning. The reason why our method outperforms various Modular Methods in terms of speed is that we do not rely on additional modules such as semantic segmentation \cite{krantz2023navigating, lei2024instanceaware}, local feature matching \cite{krantz2023navigating, lei2024instanceaware}, or switch module \cite{lei2024instanceaware} during the navigation. This simplification enables GaussNav to operate more efficiently. We also provide a qualitative example of our GaussNav navigating to the target object instance, as depicted in \Cref{fig:qualitive_result}.

\subsection{Error Analysis}

The performance of the proposed model is still far from perfect. We would like to understand the error modes for future improvements. Our analysis identifies two sources of error: the first being the model's inability to consistently match the target from instance renderings, and the second, inaccuracies in goal localization. To quantify the impact of these error sources, we conduct an evaluation of our model using a ground truth Match module and an accurate goal localization. The first one means agent can correctly recognize the target from candidate observations, and the second suggests agent is directly provided with the ground truth goal position. The variant equipped with a ground truth Match module (\textbf{GaussNav w. GT Match} in \Cref{tab:ablations}) shows that Success can be enhanced by approximately 0.127 (rows 5 and 6 in \Cref{tab:ablations}). Furthermore, when the model is augmented with both ground truth Match and Goal Localization, denoted as \textbf{GaussNav w. GT Goal Localization}, we observe an increase in Success from 0.850 to 0.946, as indicated in rows 6 and 7 in \Cref{tab:ablations}. Improvements in the first error source may be achievable through the development of a more robust re-identification algorithm. As for the second source, a more precise Grounding strategy could yield better results. These insights not only highlight the model's current shortcomings but also chart a course for subsequent refinement efforts.

\vspace{-3pt}

\subsection{Gaussian Construction Results}

As illustrated in \Cref{fig:vis_gauss}, we provide visualization results of our Semantic Gaussian. These examples demonstrate the effectiveness of our Semantic Gaussian representation across a diverse range of scenarios. By presenting a more extensive collection of results, we aim to showcase the robustness and applicability of our approach in handling various scene complexities and object compositions.

In \Cref{fig:gauss_results}, we present an quantitative evaluation of the rendering quality produced by our Semantic Gaussian Construction method on the HM3D validation dataset \cite{yadav2023habitat}. To align with the constraints of IIN task \cite{krantz2022instancespecific}, we divide each scene within HM3D into separate floors and restrict the agent's movement to within a single floor, as the IIN task \cite{krantz2022instancespecific} inherently ensures that both the agent's starting location and the target's position are on the same floor.

To quantitatively analyze the results in \Cref{fig:gauss_results}, we observe that the rendering results exhibit a bifurcated trend. For instance, in scenes with indices 29 and 30, the rendered images achieve a high PSNR of up to 40 and a near-zero depth rendering error. However, the rendering performance for scene 10 is suboptimal. We hypothesize that this polarized rendering quality across different scenes can be attributed to the discrepancy between the simulation and reality. This is evident in \Cref{fig:bad_render}, where some renderings from the HM3D dataset using Habitat simulator exhibit low fidelity, particularly in highly textured environments. High-quality reconstruction in such intricate settings is difficult, and utilizing suboptimal renderings as a basis for 3D environment reconstruction can further degrade the quality of the final output. In light of this, for scenes that are poorly reconstructed, we maintain consistency by using the original training views, rather than attempting to render novel views which would likely result in a diminished quality.

%% file: tables/main_table.tex
\begin{table}[tb]
  \centering
  \small
  \begin{tabular}{@{}l c c@{}}
    \toprule
    Method & Success~$\uparrow$ & SPL~$\uparrow$ \\
    \midrule
    RL Baseline \cite{krantz2022instancespecific} & 0.083 & 0.035  \\
    OVRL-v2 ImageNav \cite{yadav2023ovrl} & 0.006 & 0.002 \\
    OVRL-v2 IIN \cite{yadav2023ovrl} & 0.248 & 0.118 \\
    FGPrompt \cite{sun2024fgprompt} & 0.099 & 0.028 \\
    \midrule
    MultiON Baseline \cite{wani2020multion} & 0.066 & 0.045 \\
    MultiON Implicit \cite{marza2023multi} & 0.143 & 0.107 \\
    MultiON Camera \cite{chen2022learning} & 0.186 & 0.142 \\
    \midrule
    Mod-IIN \cite{krantz2023navigating} & 0.561 & 0.233 \\
    IEVE Mask RCNN \cite{lei2024instanceaware} & 0.684 & 0.241 \\
    IEVE InternImage \cite{lei2024instanceaware} & 0.702 & 0.252 \\
    \midrule
    Mod-IIN \cite{krantz2023navigating} (Scene Map) & 0.563 & 0.323 \\
    IEVE Mask RCNN \cite{lei2024instanceaware} (Scene Map) & 0.683 & 0.331 \\
    IEVE InternImage \cite{lei2024instanceaware} (Scene Map) & 0.705 & 0.347 \\
    \textbf{GaussNav (ours)} & \textbf{0.725} & \textbf{0.578}\\
    \bottomrule
  \end{tabular}
  \vspace{5pt}
  \caption{Performance comparison of our GaussNav with previous state-of-the-art methods on the HM3D \cite{yadav2023habitat} datasets across two different metrics: Success and SPL \cite{anderson2018evaluation}. The table is divided into four sections. The first section presents the results of end-to-end methods. The second section shows the transfer performance of MultiON-related methods on the IIN task. The third section includes the state-of-the-art methods on the IIN task. Finally, the fourth section aims to provide a fair comparison with GaussNav by replacing the episodic map used in these methods from the third section with a scene-specific map, allowing the agent to retain the map from the previous episode.}
  \vspace{-18pt}
  \label{tab:main}
\end{table}

%% file: tables/ablations.tex
\begin{table}[tb]
  \centering
  \small
  \begin{tabular}{@{}l c c@{}}
    \toprule
    Ablations & Success~$\uparrow$ & SPL~$\uparrow$ \\
    \midrule
    GaussNav & 0.725 & 0.578  \\
    GaussNav w.o. Classifier & 0.375 & 0.291 \\
    GaussNav w.o. Match & 0.444 & 0.353 \\
    GaussNav w.o. NVS & 0.716 & 0.557 \\
    GaussNav w. SIFT & 0.655 & 0.519 \\
    GaussNav w. GlueStick \cite{pautrat_suarez_2023_gluestick} & 0.723 & 0.577 \\
    GaussNav w. GT Match & 0.850 & 0.672 \\ 
    GaussNav w. GT Goal Localization & 0.946 & 0.744 \\
    \bottomrule
  \end{tabular}
  \vspace{5pt}
  \caption{Ablation study of GaussNav. We study the impact of Classifier, Match module, novel view synthesis (NVS) and different local feature extraction and matching algorithms on our GaussNav. The last two rows describe the performance of our method using ground truth match results and goal position.}  
  \vspace{-14pt}
  \label{tab:ablations}
\end{table}

%% file: tables/abla_classifier.tex
\begin{table}[tb]
    \centering
    \small 
    \renewcommand{\arraystretch}{1.2} 
    \setlength{\tabcolsep}{3pt} 
    \begin{tabular}{lccccccc}
        \hline
        Method & Metric & Ch. & So. & Pl. & Bed & Tol. & TV \\
        \hline
        \multirow{2}{*}{\textbf{w.o.} clf} & Success $\uparrow$ & \textbf{0.821} & \textbf{0.873} & 0.798 & \textbf{0.914} & \textbf{0.936} & 0.829 \\
        \cline{2-8}
        & Time (s) $\downarrow$ & 30.9 & 24.8 & 33.2 & 18.1 & 24.6 & 31.9 \\
        \hline
        \multirow{2}{*}{\textbf{w}. clf} & Success $\uparrow$ & 0.782 & 0.859 & \textbf{0.854} & 0.878 & 0.702 & \textbf{0.847} \\
        \cline{2-8}
        & Time (s) $\downarrow$ & \textbf{15.7} & \textbf{8.09} & \textbf{11.6} & \textbf{5.97} & \textbf{1.85} & \textbf{2.74}  \\
        \hline
    \end{tabular}
    \vspace{5pt}
    \caption{Results of matching success and time taken with or without classifier.(Abbreviations: clf = classifier, Ch. = Chair, So. = Sofa, Pl. = Plant, Tol. = Toilet)}
    \vspace{-24pt}
    \label{tab:abla_classifier}
\end{table}


%% file: tables/time.tex
\begin{table}[tb]
  \centering
  \small
  \vspace{6pt}
  \begin{tabular}{@{}c c c c c c@{}}
    \toprule
    Chair & Sofa & TV & Plant & Toilet & Bed \\
    \midrule
    11 & 2 & 1 & 3 & 1 & 0 \\
    \bottomrule
  \end{tabular}
  \vspace{5pt}
  \caption{Number of object instances across different categories in the first floor of scene \texttt{CrMo8WxCyVb}.} 
  \vspace{-24pt}
  \label{tab:time}
\end{table}

%% file: tables/NVS.tex
\begin{table*}[tb]
    \renewcommand{\arraystretch}{1.2} 
    \centering
    \begin{tabular}{c cc cc cc cc cc}
        \hline
        \multirow{2}{*}{Metric} & original & GT original & \multicolumn{2}{c}{horizontal} & \multicolumn{2}{c}{vertical} & \multicolumn{2}{c}{GT horizontal} & \multicolumn{2}{c}{GT vertical} \\
        \cline{2-11}
        & $n_v=1$ & $n_v=1$ & $n_v=3$ & $n_v=5$ & $n_v=3$ & $n_v=5$ & $n_v=3$ & $n_v=5$ & $n_v=3$ & $n_v=5$ \\
        \hline
        Success $\uparrow$ & 0.811 & 0.815 & 0.824 & 0.827 & 0.832 & 0.831 & 0.835 & 0.846 & 0.839 & 0.845 \\
        Time (s) $\downarrow$ & 11.5 & 11.7 & 32.1 & 53.9 & 33.1 & 54.2 & 32.7 & 54.1 & 33.4 & 55.1 \\
        \hline
    \end{tabular}
    \vspace{5pt}
    \caption{Results of the number of rendered images $n$, different directions (vertical or horizontal) and whether to use ground truth rendering's impact on the matching success.}
    \vspace{-16pt}
    \label{tab:NVS}
\end{table*}

%% file: tables/ieve_nvs.tex
\begin{table}[tb]
    \centering
    \vspace{5pt}
    \begin{tabular}{l c c c c c}
        \toprule
         Method & Ver. NVS & Hor. NVS & GT & Success~$\uparrow$ & SPL~$\uparrow$  \\
         \midrule
         IEVE~\cite{lei2024instanceaware} & - & - & - & 0.702 & 0.252 \\
         GaussNav (ours) & $\checkmark$ & $\times$ & $\times$ & 0.713 & 0.259 \\
         GaussNav (ours) & $\times$ & $\checkmark$ & $\times$ & 0.715 & 0.261 \\
         GaussNav (ours) & $\checkmark$ & $\checkmark$ & $\times$ & 0.723 & 0.265 \\
         GaussNav (ours) & $\checkmark$ & $\checkmark$ & $\checkmark$ & 0.747 & 0.289 \\
         \bottomrule
    \end{tabular}
    \vspace{-1pt}
    \caption{Performance Comparison of IEVE~\cite{lei2024instanceaware} and GaussNav. We compare GaussNav with the state-of-the-art method IEVE using different map representations. (Abbreviations: Ver. = Vertical, Hor. = Horizontal)}
    \vspace{-30pt}
    \label{tab:pre-exploration}
\end{table}

%% file: sections/conclusion.tex
\section{Conclusion}

In this work, we introduce a modular approach for visual navigation, \emph{i.e.,} Gaussian Splatting for Visual Navigation (GaussNav). 
Previous map-based methods largely focus on building 2D BEV map, which cannot represent the 3D geometry and detailed features in a scene. 
To this end, we propose a novel map representation, Semantic Gaussian, which is capable of preserving the scene's 3D geometry, semantic labels associated with each Gaussian, and intricate texture details. 
Leveraging this novel representation of map, we directly predict the position of target object depicted in the goal image, thereby transforming IIN into a more tractable PointGoal Navigation task. 
Our proposed framework achieves state-of-the-art performance, significantly enhancing SPL from 0.347 to 0.578. 
Furthermore, we analyze the error modes for our model and quantify the scope for improvement along two important dimensions (match and object grounding) in the future work.